\documentclass[11pt]{article}
\usepackage{latexsym,amsfonts,amsmath,epsfig,amssymb}
\usepackage[colorlinks,linkcolor=red,anchorcolor=blue,urlcolor=red,citecolor=blue]{hyperref}
\usepackage{graphicx,times}
\usepackage{mathrsfs}
\usepackage{caption}
\usepackage{subfigure}
\usepackage{graphicx}
\usepackage{epstopdf}
\usepackage{multirow}
\usepackage{booktabs}
\usepackage{makecell}

\usepackage{algorithm,algpseudocode}

\usepackage[numbers,sort&compress]{natbib}
\graphicspath{{Figuressparse/}}                  % �������е�.eps/.png �ļ���TTfigures ��Ŀ¼��

\newcommand{\qed}{\hfill $\Box$}

\newtheorem{theorem}{Theorem}[section]

\newtheorem{lemma}{Lemma}[section]

\newtheorem{remark}{Remark}[section]

\newcounter{nextauthor}
\def\mathrm{\mbox}

\numberwithin{remark}{section}

\begin{document}
\title{\Large {\bf Noisy  Nonnegative Tucker Decomposition with Sparse Factors and Missing Data}}

\author{Xiongjun Zhang\footnotemark[1]\ \ \ and \
Michael K. Ng\footnotemark[2]
}

\renewcommand{\thefootnote}{\fnsymbol{footnote}}

\footnotetext[1]{School of Mathematics and Statistics and Hubei Key Laboratory of Mathematical Sciences,
	Central China Normal University, Wuhan 430079, China (e-mail: xjzhang@ccnu.edu.cn).
	The research of this author was supported
in part by the
	National Natural Science Foundation of China under Grant No. 12171189  and the Fundamental
	Research Funds for the Central Universities under Grant No. CCNU24ai002.
}
\footnotetext[2]{Department of Mathematics, Hong Kong Baptist University,
	Kowloon Tong, Hong Kong (e-mail: michael-ng@hkbu.edu.hk).
The research of this author was supported
in part by the
Hong Kong Research Grant Council GRF 12300218, 12300519, 17201020,
17300021, C1013-21GF, C7004-21GF and Joint NSFC-RGC N-HKU76921.}

\renewcommand{\thefootnote}{\arabic{footnote}}

%\date{}
\maketitle \vspace*{0mm}
\begin{center}
\begin{minipage}{5.5in}
$${\bf Abstract}$$
Tensor decomposition is a powerful tool for extracting
physically meaningful latent factors  from multi-dimensional
nonnegative data, and has been an increasing interest
in a variety of fields such as image processing, machine learning, and computer vision.
In this paper, we propose a sparse nonnegative Tucker decomposition
and completion approach for the recovery of underlying nonnegative
data under incompleted and generally noisy observations. 
Here the underlying nonnegative
tensor data  is decomposed into a core tensor and several factor matrices
with all entries being nonnegative and the factor matrices being sparse.
The loss function is derived by the maximum likelihood estimation of the noisy observations,
and the $\ell_0$ norm is employed to enhance the sparsity of the factor matrices.
We establish the error bound of the estimator of the proposed model
under generic noise scenarios,
which is then specified to the observations with additive Gaussian noise,
additive Laplace noise, and Poisson observations, respectively.
Our theoretical results are better than those by
existing tensor-based or matrix-based methods.
Moreover, the minimax lower bounds are shown to be matched
with the derived upper bounds up to logarithmic factors.
Numerical experiments on both synthetic data and real-world data sets
demonstrate the superiority of the proposed method for nonnegative tensor
data completion.
\end{minipage}
\end{center}

\begin{center}
\begin{minipage}{5.5in}

\end{minipage}
\end{center}

\section{Introduction}
Tensors, also called multi-dimensional data, are high-order generalizations of vectors and matrices,
and have a variety of applications including  signal processing,
machine learning, computer vision, and so on \cite{kolda2009tensor, song2020robust}.
The data required from
real-world applications are usually represented as high-order tensors,
for example, color images are third-order tensors, color videos are fourth-order tensors.
In order to explore the intrinsic structure of tensor data,
tensor decomposition is a  powerful and effective approach to represent the data,
and  has been widely used in several fields
\cite{chen2018destriping, sidiropoulos2017tensor}.
Besides, tensor decomposition can identify the
fine details of tensor data and extract meaningful and explanatory information
by a lower-dimensional set of latent factors,
which is due to the fact that many high-dimensional data  reside
in a low-dimensional  subspace \cite{cichocki2009nonnegative}.

Tensor data and latent factors are
naturally nonnegative in several real-world applications such as images, video volumes, and text data.
Nonnegative tensor decomposition can be employed to extract meaningful patterns by the decomposition formulation \cite{7460200}.
For example, many high-dimensional data, such as nonnegative hyperspectral images  or video images,
are factorized to find meaningful
latent nonnegative components \cite{pan2021orthogonal}.
Besides dimension reduction,
nonnegative tensor decomposition can be  modeled and interpreted better by means of nonnegative and
sparse components, which can achieve a unique additive parts-based representation.

In particular, when the order of a tensor is second, nonnegative tensor decomposition reduces to nonnegative matrix factorization (NMF),
which has been attracted
much attention in the past decades,
see \cite{1999Learning, gillis2020nonnegative} and references therein.
The NMF, which requires the factors of
the low-rank decomposition to be componentwise nonnegative,
is capable of giving physically meaningful and more interpretable results,
and has the ability of leaning the local parts of objects \cite{1999Learning}.
There are plenty of real-world applications for NMF.
For example,
the NMF method can
extract parts of faces, such as eyes, noses, and lips, in a series of
facial images \cite{rajapakse2004color},
identify topics in a set of documents \cite{allab2016semi},
and extract materials and
their proportions in hyperspectral images unmixing \cite{lu2012manifold}.
Moreover,
sparse NMF
was introduced further because it enhances the ability of NMF to
learn a parts-based representation and produces more easily interpretable factors.
For instance, in facial feature extraction,
sparsity leads to more localized features and identifiable solutions,
while fewer features are used to reconstruct each input image \cite{gillis2020nonnegative}.
Besides, Hoyer \cite{2004Nonnegative} proposed a sparse NMF approach with sparse constraints
 to improve the decomposition for the observations with additive Gaussian noise,
and designed  a projected gradient descent algorithm to solve the resulting model.
In theory, Soni et al. \cite{soni2016noisy} proposed a noisy matrix completion
method under a class of general noise models,
and established the error bound of the estimator of their proposed model,
which can reduce to sparse NMF and completion under nonnegative constraints.
Moreover,  Sambasivan et al. \cite{Sambasivan2017Minimax}
showed that the error bound in \cite{soni2016noisy} achieves minimax error rates up to multiplicative constants and logarithmic factors.
However, for multi-dimensional data, the matrix based method may
destroy the structure of a tensor via unfolding the tensor into a matrix.

For nonnegative tensor decomposition, the key issue is the decomposition formulation of a tensor.
There are some popular decompositions for tensors  such as CANDECOMP/PARAFAC (CP) decomposition \cite{hitchcock1927expression},
Tucker decomposition \cite{Tucker1966}, tensor train decomposition \cite{oseledets2011tensor},
tensor decomposition via tensor-tensor product \cite{Kilmer2011Factorization}.
Due to the nonnegativity of a tensor,
the nonnegative CP decomposition provides an interpretable,
low tensor rank representation of the data and has been used in a variety of applications related to
 sparse image representation and image processing \cite{chen2022unsupervised}.
For example, Zhang et al. \cite{zhang2008tensor} employed a nonnegative CP decomposition for hyperspectral
unmixing with the goal to identify materials and material abundance in the data and
to compress the data.
Moreover, Chi et al. \cite{chi2012tensors} developed a Gauss-Seidel type algorithm for nonnegative CP decomposition,
where the observed data were modeled by a Poisson distribution.
 Hong et al. \cite{hong2020generalized} proposed
a generalized CP low rank tensor decomposition that
allows other loss functions besides squared error,
and presented a gradient descent type algorithm to solve the resulting model,
which can handle missing data by a similar approach in \cite{acar2011scalable}.
Besides, Jain et al. \cite{jain2017noisy} proposed a CP decomposition and completion method with
one sparse factor under a general class of noise models for third-order tensors,
and established the error bound of the estimator of their proposed model,
which was further derived to the error bound for the observations with additive Gaussian noise.
The method in \cite{jain2017noisy} can reduce to nonnegative CP decomposition
with one sparse factor under nonnegative constraint of factor matrices.
However, the previous methods need to give the CP rank of a tensor in advance,
which  is NP-hard in general \cite{hillar2013most}.

Based on the algebra framework of tensor-tensor product in \cite{Kilmer2011Factorization},
Soltani et al. \cite{soltani2016tensor} presented a
 nonnegative tensor decomposition method with one sparse factor for third-order tensors under additive Gaussian noise,
which was constructed by a tensor dictionary prior
from  training data for tomographic image reconstruction.
Furthermore, Newman et al. \cite{newman2019non} proposed
a sparse nonnegative tensor patch-based dictionary
approach according to tensor-tensor product for image compression and deblurring,
where the observations were corrupted by additive Gaussian noise.
Recently, Zhang \cite{sparse2020zhang} proposed a
sparse nonnegative tensor factorization and completion
based on the algebra framework of tensor-tensor product for third-order tensors,
where one factor tensor was sparse and the observations were corrupted by a general class of noise models.
Then the error bound of the estimator in \cite{sparse2020zhang} was established,
and the minimax lower bound was derived,
which matched with the established upper bounds
up to a logarithmic factor of the sizes of the underlying tensor.
However,
for $n$-dimensional $m$-th order tensors, there are $n^{m-2}$
$n$-by-$n$ matrix singular value decompositions to be computed \cite{Kilmer2011Factorization}
and its computational cost may be quite high in the tubular singular value decomposition
approach.
For nonnegative TT decomposition,
Lee et al. \cite{lee2016nonnegative} proposed a hierarchical alternating
least squares algorithm for feature extraction and clustering.
Although the performance of nonnegative TT decomposition was better than that of standard nonnegative Tucker decomposition,
the TT decomposition typically lacks interpretability and the efficient implementation depends on the tensor already
being in the  TT format \cite{newman2019non}.

Tucker decomposition is to decompose a given tensor into the product of a core tensor with smaller dimensions
and a series of factor matrices.
And the best low Tucker rank approximation of a tensor
was discussed and studied in \cite{2000deA}.
Due to the nonnegativity of tensor data, nonnegative Tucker decomposition (NTD)  provides an additional
multiway structured representation of tensor
data,  where all entries of the core tensor and factor matrices are nonnegative in Tucker decomposition \cite{pan2021orthogonal,4270403}.
And NTD has a wide range of applications including image denoising  and hyperspectral
image restoration \cite{7460200, bai2018nonlocal, zhou2015efficient, kim2008nonnegative}.
Moreover, M{\o}rup et al. \cite{morup2008algorithms} proposed
two multiplicative update algorithms for sparse NTD
with the observations corrupted by additive Gaussian noise and Poisson noise, respectively,
which  yields a parts-based
representation and is a more interpretable decomposition as compared to NTD.
Then Liu et al. \cite{liu2012sparse} presented a
fast and flexible algorithm for sparse
NTD with a special core tensor based on columnwise coordinate descent,
where the observations were corrupted by additive Gaussian noise
and the factor matrices were nonnegative and sparse.
Besides, Xu \cite{xu2015alternating} developed an alternating proximal gradient algorithm for
sparse NTD and completion with global convergence guarantee,
where  the observations were corrupted by additive Gaussian noise.
Zhou et al. \cite{zhou2015efficient} developed
a family of efficient first-order algorithms for sparse NTD by making use of low rank approximation,
which can reduce the storage complexity and computing time.
However,  there is no theoretical
result about the error bounds of the models
for sparse NTD with missing values in the previous work.
Also a general class of noise models is not discussed and studied
for sparse NTD and completion in the existing literature.

In this paper, we propose a sparse NTD and completion approach
for the observations with a general class of noise models.
The loss function is derived by the maximum likelihood estimation of observations,
and  the nonnegativity for the entries of the core tensor and the factor matrices in Tucker decomposition are imposed,
where the $\ell_0$ norm is used to characterise the sparsity of the factor matrices.
Moreover, the error bound of the estimator of the proposed model
under general noise  distributions is established, and then the error bounds
of the estimators for the special  observations including  additive Gaussian noise,
additive Laplace noise, and Poisson observations are derived, respectively.
Besides, the minimax lower bound of a general class of noisy observations is established,
which matches to the upper bound up to a logarithmic  factor.
Then an alternating direction method of multipliers (ADMM) is designed to solve
the resulting model. Numerical experiments on synthetic data and image data sets
demonstrate the effectiveness of the proposed model
compared with the matrix based method \cite{soni2016noisy}
and the sparse nonnegative tensor factorization and completion via tensor-tensor product \cite{sparse2020zhang}.
We summarize the existing literature
and our method for sparse nonnegative tensor decomposition/factorization in Table \ref{DiffMedSNTDF}.

\begin{table}[htbp]\scriptsize
	\begin{center}
		\setlength{\abovecaptionskip}{-1pt}
		\setlength{\belowcaptionskip}{-1pt}
		\caption{Comparisons of error bounds for sparse NTD and completion  with existing methods
based on different decompositions.}\label{DiffMedSNTDF}
		\medskip
		\begin{tabular}{| c || c|c| c| c   |} \hline
{Methods}                               & Rank              & Order & {Noise type}   & {Error bounds}       \\
			 \hline \hline
Matrix based method \cite{soni2016noisy} & matrix rank $r$   &  second-order           & general     & $O(\frac{rn_1 +\|A^*\|_0}{m}\log(\max\{n_1, n_2\}))$  \\ \hline
		     Sparse nonnegative CP \cite{jain2017noisy}         & CP rank $r$  & third-order     & general     & $O(\frac{(n_1+n_2)r+\|C\|_0}{m}\log(\max\{n_1,n_2,n_3,r\}))$   \\ \hline
		 Nonnnegative TT  \cite{lee2016nonnegative}   & tensor train rank  & \makecell{$d$th-order \\ ($d\geq 3$)} & Gaussian           & N/A  \\ \hline
%		Sparse NTF \cite{newman2019non}             & tubal rank $r$   & Gaussian       & No    & N/A   \\
		 Sparse NTF \cite{sparse2020zhang}                                   & tubal rank $r$  & third-order & general     & $O(\frac{rn_1n_3+\|\mathcal{B}\|_0}{m}\log(\max\{n_1,n_2\}))$   \\ \hline
		 \makecell{Sparse NTD \\(this paper)}          & \makecell{Tucker rank\\ $(r_1,\ldots, r_d)$ }  & \makecell{$d$th-order \\ ($d\geq 3$)}   & general          & $O(\frac{r_1\cdots r_d+\sum_{i=1}^d\|A_i\|_0}{m}\log(\max\{n_1,\ldots, n_d\}))$  \\ \hline
		\end{tabular}
	\end{center}
\end{table}

The main
	contributions of this paper are summarized as  follows.
	
(1) We propose a sparse NTD
and completion approach for the recovery of underlying nonnegative tensor 
data under incompleted and generally noisy observations, where the underlying tensor has the Tucker decomposition form  with nonnegativity and sparsity constraints.
Moreover, the generic loss function is derived by the maximum likelihood estimation of the noisy observations.

(2) An error bound of the estimator of the proposed model is established for the general observation model, which is
specified to some widely used noise models
including additive Gaussian noise, additive Laplace noise, and Poisson observations.
Moreover, the minimax lower bound of the observed model is also derived, which matches to the upper bound up to a logarithmic factor.

(3) An ADMM based algorithm is developed to solve the resulting model.
Numerical experiments on synthetic data and real-world data sets show the superior performance of the proposed method over other comparison methods for sparse NTD and completion.

The remaining parts of this paper are organized as follows. In Section \ref{Prelim}, some
preliminaries about tensor Tucker decomposition and related information theoretical quality are provided.
In Section \ref{ProMod}, a sparse NTD and completion model is proposed under a general class of noisy observations.
An upper error bound of the estimator of the proposed model is established in Section \ref{upperbound}.
Moreover, the error bounds of the estimators are further derived for the observations with special noise models.
Section \ref{lowerbou} is devoted to the minimax lower bound of the estimator.
In Section \ref{Algorithm}, an ADMM based algorithm is developed to solve the resulting model.
In Section \ref{NumerExper}, numerical experiments are conducted  to demonstrate the effectiveness of the proposed model.
Discussions and future work are given in Section \ref{Conclu}.
All proofs of theorems are provided in the appendices.

\section{Preliminaries}\label{Prelim}

\subsection{Notation}

Let  $\mathbb{R}^n$ and $\mathbb{R}_+^{n_1\times n_2\times\cdots \times n_d}$
denote the $n$-dimensional Euclidean space with real numbers
and the set of $n_1\times n_2\times\cdots \times n_d$ tensors with nonnegative real entries, respectively.
Scalars, vectors, matrices, and tensors are represented by lowercase letters,
lowercase boldface letters, uppercase letters, and capital Euler script letters, respectively,
e.g., $x, \mathbf{x}, X$, $\mathcal{X}$.
$\log$ refers to natural logarithm throughout this paper.
$\lfloor x\rfloor$ and $\lceil x\rceil$ are the integer part of $x$
and  smallest integer that is larger or equal to $x$, respectively.
We denote $n_m=\max\{n_1,n_2,\ldots, n_d\}$. 
%$a\vee b=\max\{a,b\}$, and $a\wedge b=\min\{a,b\}$, respectively.
For any positive
integer $n$, we let $[n]=\{1,2,\ldots, n\}$.
For an arbitrary set $\mathfrak{C}$, $|\mathfrak{C}|$ denotes the number of entries in $\mathfrak{C}$.

Let $\mathcal{X}_{i_1i_2\cdots i_d}$ denote the $(i_1,i_2,\ldots, i_d)$th entry of a tensor $\mathcal{X}\in\mathbb{R}^{n_1\times n_2\times \cdots \times  n_d}$.
 The number of ways or dimensions of a tensor is called the order \cite{kolda2009tensor}.
For an arbitrary vector $\mathbf{x}\in\mathbb{R}^{n}$,
$\|\mathbf{x}\|$ and $\|\mathbf{x}\|_1$ denote the Euclidean norm and  the $\ell_1$ norm of $\mathbf{x}$, respectively,
where $\|\mathbf{x}\|_1=\sum_{i=1}^n|x_i|$ and $x_i$ is the $i$th component of $\mathbf{x}$, $i\in[n]$.
$\|X\|_0$ denotes the number of nonzero entries of a matrix $X$.
The tensor Frobenius norm of $\mathcal{X}\in\mathbb{R}^{n_1\times n_2\times  \cdots \times n_d}$ is defined as $\|\mathcal{X}\|_F=\sqrt{\langle \mathcal{X}, \mathcal{X}\rangle}$,
where the inner product of two  same-sized tensors
is defined as $\langle \mathcal{X},\mathcal{Y} \rangle
=\sum_{i_1,i_2,\ldots, i_d}\mathcal{X}_{i_1i_2\cdots i_d}\mathcal{Y}_{i_1i_2\cdots i_d}$.
The tensor infinity norm of $\mathcal{X}$, denoted by $\|\mathcal{X}\|_\infty$, is defined as
$\|\mathcal{X}\|_\infty=\max_{i_1,i_2,\ldots,i_d}|\mathcal{X}_{i_1i_2\cdots i_d}|$.

The mode-$i$ unfolding of a tensor $\mathcal{X}\in\mathbb{R}^{n_1\times n_2\times  \cdots\times n_d}$,
 denoted by $\mathcal{X}_{(i)}$, arranges the mode-$n$ fibers to be the columns of the resulting matrix,
 where the $(i_1,i_2,\ldots, i_d)$th entry of $\mathcal{X}$ maps to matrix entry $(i_n,j)$ with
 $$
j=1+\sum_{k=1 \atop k\neq n}^d(i_k-1)J_k \ \textup{and} \ J_k=\prod_{l=1 \atop l\neq n}^{k-1}n_l.
$$
Here a fiber is defined by fixing every index but one.
The vectorization of a tensor  $\mathcal{X}$ is a vector, which is obtained by stacking all mode-$1$
fibers of $\mathcal{X}$ and denoted by vec($\mathcal{X}$).

Let $\mathfrak{R}$ be a Euclidean space endowed with the Euclidean norm $\|\cdot\|$.
For an arbitrary closed proper function $g:\mathfrak{R}\rightarrow (-\infty,+\infty]$,
the proximal mapping of $g$ at $y$ is the operator given by
$$
\mbox{Prox}_g(y)=\arg\min_{x\in\mathfrak{R}}\left\{g(x)+\frac{1}{2}\|x-y\|^2\right\}.
$$

\subsection{Multilinear Operators}

The Tucker decomposition of a $d$th-order tensor $\mathcal{X}\in\mathbb{R}^{n_1\times n_2\times  \cdots \times n_d}$ is given by \cite{Tucker1966}
$$
\mathcal{X}=\mathcal{C}\times_1A_1\times_2\cdots\times_d A_d,
$$
where $\mathcal{C}\in\mathbb{R}^{r_1\times r_2\times  \cdots\times  r_d}$ is called the core tensor,
$A_i\in\mathbb{R}^{n_i\times r_i}$ are the factor matrices, $i\in[d]$,
and $(r_1,r_2,\ldots, r_d)$ is called the Tucker rank of $\mathcal{X}$.
Here the $n$-mode product of a tensor $\mathcal{C}$ by a matrix $A\in\mathbb{R}^{n\times r_i}$,
denoted by $\mathcal{C}\times_i A$, is a
$r_1\times\cdots \times r_{i-1}\times n\times r_{i+1}\times \cdots\times r_d$-tensor
whose entries are given by \cite[Definition 3]{2000deA}
$$
(\mathcal{C}\times_i A)_{j_1j_2\cdots j_{i-1}k_ij_{i+1}\cdots j_d}=
\sum_{j_i=1}^{r_i}\mathcal{C}_{j_1j_2\cdots j_{i-1}j_ij_{i+1}\cdots j_d}A_{k_ij_i}.
$$
It is easy to verify that if $\mathcal{X}=\mathcal{C}\times_1A_1\times_2\cdots\times_d A_d$, then
\begin{equation}\label{VextTen}
\textup{vec}(\mathcal{X})=(\otimes_{i=d}^1A_i)\textup{vec}(\mathcal{C}),
\end{equation}
where
$
\otimes_{i=d}^1A_i:=A_d\otimes A_{d-1}\otimes\cdots\otimes A_1
$
and $A\otimes B$ denotes the Kronecker product of $A$ and $B$. Moreover, the mode-$n$ unfolding of $\mathcal{X}$ is given by \cite{kolda2009tensor}
\begin{equation}\label{TenUnfol}
\mathcal{X}_{(n)} = A_{n}\mathcal{C}_{(n)}(\otimes_{i=d, i\neq n}^1A_i)^T.
\end{equation}

\subsection{Kullback-Leibler Divergence and Hellinger Affinity}

Let $p_{x_1}(y)$ and $p_{x_2}(y)$ denote the
probability density functions or probability mass functions
of a real scalar random variable $y$ with corresponding parameters $x_1$ and $x_2$, respectively.
The Kullback-Leibler (KL) divergence of $p_{x_1}(y)$ from $p_{x_2}(y)$ is denoted and defined as follows:
$$
K(p_{x_1}(y)||p_{x_2}(y))=\mathbb{E}_{p_{x_1}(y)}\left[\log\frac{p_{x_1}(y)}{p_{x_2}(y)}\right].
$$
The Hellinger affinity between $p_{x_1}(y)$ and $p_{x_2}(y)$   is denoted and defined as
$$
H(p_{x_1}(y)||p_{x_2}(y))
=\mathbb{E}_{p_{x_1}}\left[\sqrt{\frac{p_{x_2}(y)}{p_{x_1}(y)}}\right]
=\mathbb{E}_{p_{x_2}}\left[\sqrt{\frac{p_{x_1}(y)}{p_{x_2}(y)}}\right].
$$
The joint probability distributions of random tensors, denoted
by $p_{\mathcal{X}_1}(\mathcal{Y}), p_{\mathcal{X}_2}(\mathcal{Y})$,
are the joint probability distributions of the vectorization of the tensors.
Then the KL divergence of $p_{\mathcal{X}_1}(\mathcal{Y})$ from $ p_{\mathcal{X}_2}(\mathcal{Y})$ is defined as
$$
K(p_{\mathcal{X}_1}(\mathcal{Y})||p_{\mathcal{X}_2}(\mathcal{Y}))
:=\sum_{i_1,i_2,\ldots, i_d}K(p_{(\mathcal{X}_1)_{i_1i_2\cdots i_d}}(\mathcal{Y}_{i_1i_2\cdots i_d})||p_{(\mathcal{X}_2)_{i_1i_2\cdots i_d}}(\mathcal{Y}_{i_1i_2\cdots i_d})),
$$
and its Hellinger affinity is defined as
$$
H(p_{\mathcal{X}_1}(\mathcal{Y})||p_{\mathcal{X}_2}(\mathcal{Y}))
:=\prod_{i_1,i_2,\ldots, i_d}H(p_{(\mathcal{X}_1)_{i_1i_2\cdots i_d}}(\mathcal{Y}_{i_1i_2\cdots i_d})|| p_{(\mathcal{X}_2)_{i_1i_2\cdots i_d}}(\mathcal{Y}_{i_1i_2\cdots i_d})).
$$

\section{Sparse NTD and Completion With Noisy Observations}\label{ProMod}

Let  $\mathcal{X}^*\in\mathbb{R}_+^{n_1\times n_2\times \cdots \times n_d}$ be an unknown nonnegative tensor with Tucker rank $(r_1,r_2,\ldots, r_d)$.
Assume that $\mathcal{X}^*$ admits a following sparse nonnegative Tucker decomposition:
\begin{equation}\label{TucDeco}
\mathcal{X}^*=\mathcal{C}^*	\times_1 A_1^*\times_2 A_2^*\cdots \times_d A_d^*,
\end{equation}
where $\mathcal{C}^*\in\mathbb{R}_+^{r_1\times r_2\times \cdots \times r_d}$ is the core tensor and
 $A_i^*\in\mathbb{R}_+^{n_i\times r_i}$
 are sparse factor matrices, $i\in[d]$. Suppose that $r_i\leq n_i, i\in[d]$.
 We also assume that each entries of $\mathcal{X}^*, \mathcal{C}^*,
A_i^*$ are bounded, i.e.,
\begin{equation}\label{Constecf}
\begin{split}
&0\leq \mathcal{X}_{i_1i_2\cdots i_d}^*\leq \frac{c}{2}, \ \  (i_1,i_2,\ldots,i_d)\in[n_1]\times [n_2]\times \cdots \times [n_d],  \\
& 0\leq \mathcal{C}_{i_1i_2\cdots i_d}^*\leq 1, \ \ (i_1,i_2,\ldots,i_d)\in[r_1]\times [r_2]\times \cdots \times [r_d], \\
& 0\leq (A_{i}^*)_{lm}\leq a_i, \  \ (l,m)\in[n_i]\times[r_i],  i\in[d],
\end{split}
\end{equation}
where $c, a_i>0$ are given constants.
Here we use $\frac{c}{2}$ for brevity in the subsequence analysis.

However,
only a noisy and incompleted  version of the underlying tensor $\mathcal{X}^*$ is available in practice.
Let $\Omega\subseteq[n_1]\times [n_2]\times \cdots\times  [n_d]$
be a subset at which the entries of the observations $\mathcal{Y}$ are collected.
Denote $\mathcal{Y}_{\Omega}\in\mathbb{R}^m$ to be a vector
such that the entries of $\mathcal{Y}$ in the index set $\Omega$ are vectorized  by lexicographic order,
where $m$ is the number of the observed entries.
Assume that $n_1, n_2, \ldots, n_d\geq 2$ and $d\geq 3$ throughout this paper.
Let the location
$\Omega$ set be generated according to an independent
Bernoulli model with probability $p=\frac{m}{n_1n_2\cdots n_d}$ (denoted by Bern($p$)),
i.e., each index $(i_1,i_2,\ldots, i_d)$ belongs to $\Omega$ with probability $p$,
which is denoted as $\Omega\sim  \text{Bern}(p)$.
Suppose that the entries of observations are conditionally independent,
which implies that the overall likelihood is just the product of the likelihoods of each entry.
Therefore, the joint probability density function
or probability mass function of observations $\mathcal{Y}_\Omega$ is given by
\begin{equation}\label{obserPo}
p_{\mathcal{X}_\Omega^*}(\mathcal{Y}_{\Omega})
:=\prod_{(i_1,i_2,\ldots, i_d)\in \Omega}p_{\mathcal{X}_{i_1i_2\cdots i_d}^*}(\mathcal{Y}_{i_1i_2\cdots i_d}).
\end{equation}

By taking the negative logarithm of the probability distribution,
we propose the following sparse NTD and completion model
with nonnegative and bounded constraints:
\begin{equation}\label{model}
\min_{\mathcal{X}\in \Upsilon}\left\{-\log p_{\mathcal{X}_\Omega}(\mathcal{Y}_{\Omega})+\sum_{i=1}^d\lambda_i\|A_i\|_0\right\},
\end{equation}
where $\lambda_i>0$ are the regularization parameters, $i\in[d]$,
$\|A_i\|_0$ is employed to characterize the sparsity of the factor matrix $A_i$ in Tucker decomposition,
and $\Upsilon$ is defined as
\begin{equation}\label{TauSet}
\begin{split}
\Upsilon:=\Big\{ & \mathcal{X}=\mathcal{C}	\times_1A_1\cdots\times_d A_d:
\ \mathcal{C}\in\mathfrak{C}, \ A_i\in\mathfrak{B}_i, i\in [d], \\
& 0\leq \mathcal{X}_{i_1i_2\cdots i_d}\leq c, \  (i_1,i_2,\ldots,i_d)\in[n_1]\times [n_2]\times \cdots \times [n_d] \Big\}.
\end{split}
\end{equation}
Here $\Upsilon$ is a countable set of candidate estimates,
and $\mathfrak{C}$ and $\mathfrak{B}_i$ are the nonnegative and bounded sets
constructed as follows:
First, let
\begin{equation}\label{denu}
\tau:=2^{\lceil\log_2(n_m)^\beta\rceil}
\end{equation}
for a specified $\beta\geq 1.$
Then, we construct $\mathfrak{C}$ to be the set
of all nonnegative tensors $\mathcal{C}\in\mathbb{R}_+^{r_1\times r_2\times \cdots \times r_d}$
whose entries are discretized to one of $\tau$
uniformly sized bins in the range $[0,1]$,
and $\mathfrak{B}_i$
to be the set of all nonnegative matrices $A_i\in\mathbb{R}_+^{n_i\times r_i}$
whose entries either take the value $0$, or are discretized to
one of $\tau$ uniformly sized bins in the range $[0,a_i]$, $i\in[d]$.

\begin{remark}
The model in (\ref{model}) can handle the observations with general noise distributions,
where one just needs to know the probability mass function
or probability density function of the  observations $\mathcal{Y}_\Omega$.
In particular,
the noise models including additive Gaussian noise,
additive Laplace noise, and Poisson observations
will be discussed in detail in the next section.
\end{remark}

\begin{remark}
For the observations with additive Gaussian noise,
Liu et al. \cite{liu2012sparse} proposed a sparse NTD model with special core tensor $\mathcal{C}$, which is effective for image compression. Moreover, Xu \cite{xu2015alternating}
proposed a gradient descent type algorithm for sparse NTD and completion with additive Gaussian noise,
which used the relaxation of $\ell_0$ norm for each factor matrix.
Besides, M{\o}rup et al.  \cite{morup2008algorithms} proposed two multiplicative update algorithms
for sparse NTD, where additive Gaussian noise and Poisson observations were considered with full observations.
However, there is no theoretical analysis of error
 bounds about these models in \cite{liu2012sparse,morup2008algorithms, xu2015alternating}.
Only efficient algorithms were proposed and studied to solve their resulting sparse NTD models.
Moreover, a unified  framework for general loss functions  is proposed in (\ref{model}) for sparse NTD and completion,
while the special noise type is considered in the existing literature,
such as additive Gaussian noise \cite{liu2012sparse, xu2015alternating}, Poisson observations  \cite{morup2008algorithms}.
\end{remark}

\begin{remark}
Recently,  Zhang et al. \cite{sparse2020zhang} proposed a
sparse nonnegative tensor factorization and completion model
with general noise distributions based on the algebra framework of tensor-tensor
product for third-order tensors,
where the underlying tensor was decomposed
into the tensor-tensor product of  one sparse nonnegative tensor and one nonnegative tensor.
The difference between model (\ref{model}) and \cite{sparse2020zhang} is
the factorization of the underlying tensor, where the tensor-tensor
product with block circulant structure 
is used in \cite{sparse2020zhang} and the Tucker
decomposition of the underlying tensor is utilized in model (\ref{model}).
And the nonnegative Tucker decomposition has more
widely applications and explanatory information than the nonnegative
tensor factorization based on tensor-tensor product,
 where the last one is mainly applied in X-ray CT imaging,
 image compression and deblurring based on  tensor patch dictionary learning \cite{newman2019non}.
Besides, the model in \cite{sparse2020zhang} is only effective for third-order tensors,
 while the model in (\ref{model}) can be employed to any order tensors
 and can extract more physically meaningful latent components by the sparse and nonnegative factors \cite{zhou2015efficient}.
\end{remark}

\begin{remark}
Jain  et al. \cite{jain2017noisy} proposed a noisy tensor completion method based on CP decomposition,
where one factor is sparse.
The model in (\ref{model}) can reduce to sparse nonnegative
CP decomposition and completion when the core tensor is
diagonal.
However, the core tensor of Tucker decomposition is not diagonal in general
 in various real-world applications \cite{zhou2015efficient}.
 Moreover, Hong et al. \cite{hong2020generalized} proposed a generalized CP decomposition with  a general loss
 function under various scenarios
  and designed a gradient descent algorithm to solve the resulting model,
  while there are no nonnegative and sparse constraints for the factor matrices, and
  there is no theoretical analysis of error bounds
about the model in \cite{hong2020generalized}.
\end{remark}

\begin{remark}
In model (\ref{model}), we only require that the factor matrices are sparse and do not impose 
additional sparsity constraints on the core tensor, which has been applied in feature extraction \cite{zhou2015efficient} and high-dimensional time series \cite{wang2022high}.
Furthermore, our model can be generalized to the model with the core tensor and factor matrices being sparse simultaneously.
\end{remark}

\section{Error Bounds}\label{upperbound}

In this section, an upper error
bound of the estimator of the sparse NTD and completion model in (\ref{model}) is established under a general class of noise distributions,
and then the upper error bounds of the estimators for the observations with special noise models are derived,
including additive Gaussian noise, additive Laplace noise, and Poisson observations.

Let $\mathcal{X}^{\lambda}$ be a minimizer of (\ref{model}).
Next we state the main theorem about the error bound of the estimator in (\ref{model}).

\begin{theorem}\label{mainthe}
Let the sampling set $\Omega$ be drawn from the independent
Bernoulli model with probability $p=\frac{m}{n_1n_2\cdots n_d}$, i.e., $\Omega\sim \textup{Bern}(p)$,
and the joint probability density/mass function  of $\mathcal{Y}_\Omega$ be defined  as (\ref{obserPo}).
For any
$$
\lambda_i\geq 4(\beta+2)\left(1+\frac{2\gamma}{3}\right)\log(n_m),  \ \ i\in[d],
$$
where  $\gamma$ is a constant satisfying
\begin{equation}\label{kapparThe}
\gamma\geq \max_{\mathcal{X}\in\Upsilon}\max_{i_1,i_2,\ldots, i_d} K\left(p_{\mathcal{X}_{i_1i_2\cdots i_d}^*}(\mathcal{Y}_{i_1i_2\cdots i_d})||p_{\mathcal{X}_{i_1i_2\cdots i_d}}(\mathcal{Y}_{i_1i_2\cdots i_d})\right),
\end{equation}
then
\[
\begin{split}
 & \ \frac{\mathbb{E}_{\Omega,\mathcal{Y}_{\Omega}}  \left[-2\log H(p_{{\mathcal{X}}^{\lambda}}(\mathcal{Y}) || p_{\mathcal{X}^*}(\mathcal{Y}))\right]}{n_1n_2\cdots n_d}
 \leq  \frac{8\gamma\log(m)}{m}+ 3\cdot\min_{\mathcal{X}\in\Upsilon}\Biggl\{ \frac{K(p_{\mathcal{X}^*}(\mathcal{Y})|| p_{\mathcal{X}}(\mathcal{Y}))}{n_1n_2\cdots n_d} \\
& \ ~~~~~~~~~~~~~~~~~~~~~~~~~~~~~~~~~~~~ +\left(\max_i\{\lambda_i\}+\frac{8\gamma(\beta+2)\log(n_m)}{3}\right)\frac{r_1r_2\cdots r_d+\sum_{i=1}^d\|A_i\|_0}{m} \Biggl\},
\end{split}
\]
where the expectation is taken with regard to the joint distribution of $\Omega$ and $\mathcal{Y}_{\Omega}$.
\end{theorem}

Theorem \ref{mainthe} is the upper error bound of the estimator
in (\ref{model}) under a general class of noise distributions,
which can be specified to  some special noise distributions,
such as additive Gaussian noise, additive Laplace noise, and Poisson observations.
The key point for showing the upper error bound with special noise distributions
is to specify the logarithmic Hellinger affinity $\log H(p_{{\mathcal{X}}^{\lambda}}(\mathcal{Y}),p_{\mathcal{X}^*}(\mathcal{Y}))$ and the minimum KL divergence $\min_{\mathcal{X}\in \Upsilon}K(p_{\mathcal{X}^*}(\mathcal{Y})|| p_{\mathcal{X}}(\mathcal{Y}))$.

From Theorem  \ref{mainthe}, we can see that the negative logarithmic Hellinger affinity with respect to the underlying tensor
and the recovered tensor is on the order of $O(\frac{r_1r_2\cdots r_d+\sum_{i=1}^d\|A_i\|_0}{m}\log(n_m))$
if the KL divergence  $\min_{\mathcal{X}\in \Upsilon}K(p_{\mathcal{X}^*}(\mathcal{Y})|| p_{\mathcal{X}}(\mathcal{Y}))$ is not too large
and $\lambda_i$ are fixed, where $A_i$ are the factor matrices of $\mathcal{X}\in\Upsilon$ in the Tucker decomposition.
Therefore, if the Tucker rank is low, the error bound of the estimator in (\ref{model}) will be small.

\begin{remark}
For the problem of sparse NTD with partial observations, the upper error
bound is related to the degree of freedoms of the tensor,
i.e., $r_1r_2\cdots r_d+\sum_{i=1}^d \|A_i\|_0$.
Moreover, it follows from Theorem \ref{mainthe} that the upper error bound decreases
as the number of observed samples increases.
\end{remark}

In the following subsections, we specify the noise models and establish
the detailed upper error bounds based on Theorem \ref{mainthe}.
In the discretization levels (\ref{denu}), we choose
\begin{equation}\label{betadef}
\beta = 1+ \frac{\log\left(\frac{(2^{d+1}-1)\sqrt{d}r_1r_2\cdots r_d a_1a_2\cdots a_d}{c \sqrt{n_m}}+1\right)}{\log(n_{m})},
\end{equation}
which implies that $\beta\geq 1$.
For the regularization parameters $\lambda_i$, we consider the specific choice
\begin{equation}\label{lambdai}
\lambda_i=  4(\beta+2)\left(1+\frac{2\gamma}{3}\right)\log(n_m),  \ \ i\in[d],
\end{equation}
where $\gamma$ satisfies (\ref{kapparThe}) and is related to the maximum KL divergence between
$p_{\mathcal{X}_{i_1i_2\cdots i_d}^*}(\mathcal{Y}_{i_1i_2\cdots i_d})$ and
$p_{\mathcal{X}_{i_1i_2\cdots i_d}}(\mathcal{Y}_{i_1i_2\cdots i_d})$ for any $\mathcal{X}\in\Upsilon$.

\subsection{Additive Gaussian Noise}

In this subsection, we establish the upper error bound of the estimator  in (\ref{model})
for the observations corrupted by additive Gaussian noise, which has been widely used in a variety of applications.
In this case, the observation $\mathcal{Y}_\Omega$
is generated by
$$
\mathcal{Y}_{i_1i_2\cdots i_d} = \mathcal{X}_{i_1i_2\cdots i_d}^* + \mathcal{N}_{i_1i_2\cdots i_d}, \ \ \
(i_1,i_2,\ldots, i_d)\in \Omega,
$$
where $\mathcal{N}_{i_1i_2\cdots i_d}$ are independent zero-mean Gaussian noise
with standard deviation $\sigma>0$. The distribution of
 $\mathcal{Y}_\Omega$ obeys a multivariate Gaussian density with dimension $|\Omega|$
 whose mean and  covariance matrix are $\mathcal{X}^*_\Omega$ and $\sigma^2 I_{\Omega}$, respectively,
where $I_{\Omega}$ denotes the identity matrix with dimension $|\Omega|\times |\Omega|$.
Then the joint probability density function of $\mathcal{Y}_\Omega$ is given by
\begin{equation}\label{Gausspdf}
p_{\mathcal{X}_\Omega^*}(\mathcal{Y}_{\Omega})
=\frac{1}{(2\pi\sigma^2)^{|\Omega|/2}}\exp\left(-\frac{\|\mathcal{Y}_\Omega-\mathcal{X}_\Omega^*\|^2}{2\sigma^2}\right).
\end{equation}

Now we specify the upper error bound in Theorem \ref{mainthe}
for the observations with  additive Gaussian noise in the following theorem,
where the joint probability density function of the observations is given by (\ref{Gausspdf}).

\begin{theorem}\label{errspeGaNo}
Let $\beta$ and $\lambda_i$ be defined as (\ref{betadef}) and (\ref{lambdai}), $i\in[d]$,
where $\gamma=\frac{c^2}{2\sigma^2}$ in  (\ref{lambdai}). Assume that the sampling
set $\Omega\sim \textup{Bern}(p)$ with $p=\frac{m}{n_1n_2\cdots n_d}$ and the joint
probability density function of $\mathcal{Y}_\Omega$ is given by
(\ref{Gausspdf}).
Then the estimator in (\ref{model}) satisfies
\[
\begin{split}
  \frac{\mathbb{E}_{\Omega,\mathcal{Y}_{\Omega}}\left[\|\mathcal{X}^\lambda-\mathcal{X}^*\|_F^2\right]}{n_1n_2\cdots n_d}
\leq & \ \frac{22c^2\log(m)}{m} \\
&  \ + 16(\beta+2)(2c^2+3\sigma^2)\left(\frac{r_1r_2\cdots r_d+\sum_{i=1}^d\|A_i^*\|_0}{m}\right)\log(n_m).
\end{split}
\]
\end{theorem}

By Theorem \ref{errspeGaNo}, we know that the upper error bound of the estimator
of (\ref{model}) for the observations with additive Gaussian noise is on the order of
$$
O\left(\frac{r_1r_2\cdots r_d+\sum_{i=1}^d\|A_i^*\|_0}{m}\log(n_m)\right).
$$
Consequently, if the factor matrices are sparser, the error upper bound of the estimator is lower.
Moreover, when $m=n_1\cdots n_d$ and  $\|A_i\|_0 =r_in_i, i\in[d]$, which implies that $A_i$ are dense,
the error bound in Theorem \ref{errspeGaNo} is just the error bound of the estimator of NTD with additive Gaussian noise,
which has been widely used and investigated, see \cite{cichocki2009nonnegative,  morup2008algorithms} and references therein.

\begin{remark}
For sparse NTD with additive Gaussian noise,
Liu et al. \cite{liu2012sparse} proposed a novel model with a special core tensor in the Tucker decomposition,
where the $\ell_0$ norm is replaced by the $\ell_1$ norm for the factor matrices.
In this case,
Theorem \ref{errspeGaNo} establishes the error bound of the model  in \cite{liu2012sparse}
by employing the $\ell_0$ norm to characterize the sparsity of the factor matrices.
Moreover, Xu et al. \cite{xu2015alternating}
proposed a sparse NTD and completion approach for the observations with additive Gaussian noise,
which replaced the $\ell_0$ norm by the $\ell_1$ norm in (\ref{model}).
This implies that Theorem \ref{errspeGaNo} established
the error bound of the estimator of the model in  \cite{xu2015alternating}
when the $\ell_0$ norm is utilized to measure the sparsity of the factor matrices.
\end{remark}

\begin{remark}
If we ignore the internal structure of a tensor and unfold the tensor into a matrix,
we can compare the error bound in Theorem \ref{errspeGaNo} with that in \cite[Corollary 3]{soni2016noisy},
where the matrix based method in \cite{soni2016noisy} can
reduce to sparse nonnegative matrix factorization and completion with nonnegative and bounded constraints.
For a given tensor $\mathcal{X}^*$ in the form of (\ref{TucDeco}), without loss of generality,
we unfold it along the first mode, i.e.,
 $\mathcal{X}_{(1)}^*=A_{1}^*(\mathcal{C}_{(1)}^*(\otimes_{i=d}^2A_i^*)^T).$
In this case, one is capable of applying  the matrix based method in \cite{soni2016noisy}
to $(\mathcal{X}_{(1)}^*)^T\in\mathbb{R}^{(n_2\cdots n_d)\times n_1}_+$, where the sparse factor is $A_1^*$.
Denote $\mathcal{X}^m$ to be the estimator by the matrix method, 
then the resulting error bound of the estimator of  the matrix based method \cite[Corollary 3]{soni2016noisy} is 
\[
\begin{split}
 &\frac{\mathbb{E}_{\Omega,\mathcal{Y}_{\Omega}}\left[\|\mathcal{X}^m-\mathcal{X}^*\|_F^2\right]}{n_1n_2\cdots n_d}\\
 \leq \ & \frac{70c^2\log(m)}{m}+8(3\sigma^2+8c^2)(\beta_1+2)\log(\max\{n_1, n_2\cdots n_d\})\frac{r_1n_2\cdots n_d +\|A_1^*\|_0}{m},
\end{split}
\]
where $\beta_1=\max\{1, 1+\frac{\log(8r_1a_1/c)}{\log(\max{n_1, n_2\cdots n_d})}\}$.
This implies that the order of the error bound  of the estimator is
$
O(\frac{r_1n_2\cdots n_d +\|A_1^*\|_0}{m}\log\left(\max\{n_1, n_2\cdots n_d\}\right)).
$
Now we compare it with the case that only nonnegative tensor decomposition is considered.
In this case, the error bound of the matrix based method in  \cite{soni2016noisy}
 for the tensor with mode-$1$ unfolding is the order of
$O(\frac{n_1r_1+r_1n_2n_3\cdots n_d}{m}\log(\max\{n_1, n_2\cdots n_d\})$.
And the error bound in Theorem \ref{errspeGaNo} is  the order of $O(\frac{r_1r_2\cdots r_d+\sum_{i=1}^dr_in_i}{m}\log(n_m))$.
Therefore, the error bound in Theorem \ref{errspeGaNo} is smaller
than that in \cite{soni2016noisy} if $r_i$ is much smaller than $n_i, i\in[d]$ and $d$ is not too large.
In particular, if $r_1= \cdots =r_d=r$ and $n_1=\cdots =n_d=n$, the error bounds
in Theorem \ref{errspeGaNo} and in \cite{soni2016noisy}
are $O(\frac{r^d+drn}{m}\log(n))$ and $O(\frac{rn+rn^{d-1}}{m}\log(n(d-1))$, respectively,
where the error bound of the matrix based method is larger than that in Theorem \ref{errspeGaNo} if $r,d$ are much smaller than $n$.
\end{remark}

\begin{remark}
We compare the error bound in Theorem \ref{errspeGaNo} with that in \cite[Proposition 4.1]{sparse2020zhang},
which is based on the algebra framework of tensor-tensor product and is only effective for third-order tensors.
In general, the two error bound results are not comparable directly
since the sparse factors are different for the two factorizations.
However, when the factors matrices in Tucker decomposition
and the factor tensor in the algebra framework of tensor-tensor product are dense,
the two error bounds can be comparable. For a third-order tensor with size $n_1\times n_2 \times n_3$,
the error bound of the estimator  in Theorem \ref{errspeGaNo} is
$
O(\frac{r_1r_2r_3+\sum_{i=1}^3r_in_i}{m}\log(n_m)),
$
while it is $O(\frac{rn_1n_3+rn_2n_3}{m}\log(\max\{n_1, n_2\}))$ in \cite{sparse2020zhang},
where $r$ is the tubal rank of the underlying tensor.
In this case, the error bound in Theorem \ref{errspeGaNo} is smaller than
that in \cite{sparse2020zhang} if $r_i$ is much smaller than $n_i$
and $r$ is close to $r_i$. In particular, if $n_1=n_2=n_3=n$
and $r_1=r_2=r_3=r$, the error bound in Theorem \ref{errspeGaNo}
is
$
O(\frac{r^3+3rn}{m}\log(n)),
$
which is smaller than that (i.e., $O(\frac{2rn^2}{m}\log(n))$) in \cite{sparse2020zhang} if $r\leq 2n-3$ and $n\geq 3$.
These conditions can be satisfied easily in real-world applications since $n$ is generally large.
\end{remark}

\begin{remark}
Jain  et al. \cite{jain2017noisy} proposed a noisy tensor completion
model based on CP decomposition with a special sparse  factor
for an $n_1\times n_2\times n_3$ tensor, where the third factor matrix is sparse in CP decomposition.
In particular, when the constraints are nonnegative, the model in \cite{jain2017noisy}
reduces to sparse nonnegative CP decomposition and completion.
In this case, the error bound of the estimator of their model for the observations with additive
Gaussian noise was
\begin{equation}\label{CPEUB}
\frac{\mathbb{E}_{\Omega,\mathcal{Y}_{\Omega}}[\|\mathcal{X}_{cp}-\mathcal{X}^*\|_F^2]}{n_1n_2n_3}=O\left(\frac{(n_1+n_2)r+\|C^*\|_0}{m}\log(\max\{n_1, n_2,n_3\})\right),
\end{equation}
where $\mathcal{X}_{cp}$ is the estimator by \cite{jain2017noisy},
 $r$ is the CP rank of $\mathcal{X}^*$,
 and $C^*$ is the third factor matrix in the CP decomposition of $\mathcal{X}^*$.
 However, (\ref{CPEUB}) is hard to compare with the error bound in Theorem \ref{errspeGaNo} directly
since the CP rank of a tensor is not comparable to the Tucker rank of a tensor in general.
Also the factor matrices are typically different between the CP and Tucker decompositions of a tensor.
In particular, if the core tensor in Tucker decomposition is diagonal and $r_1=r_2=r_3=r$,
the error bound in Theorem \ref{errspeGaNo}
is comparable with (\ref{CPEUB}).
In fact, the error bound in Theorem \ref{errspeGaNo} is smaller than (\ref{CPEUB}) if $r$ is much smaller than $n_i$.
\end{remark}

\subsection{Additive Laplace Noise}

In this subsection, we establish the error bound of the estimator of model
(\ref{model}) for the observations with additive Laplace noise.
In this case, the observation $\mathcal{Y}_\Omega$
is modeled by
\begin{equation}\label{LapObN}
\mathcal{Y}_{i_1\cdots i_d} = \mathcal{X}_{i_1\cdots i_d}^* + \mathcal{N}_{i_1\cdots i_d}, \ \ \ (i_1,\ldots, i_d)\in \Omega,
\end{equation}
where $\mathcal{N}_{i_1\cdots i_d}$ are independent Laplace distribution with parameters $(0,\tau)$, $\tau>0$,
which is denoted by $\textup{Laplace}(0,\tau)$.
The joint probability density function of $\mathcal{Y}_\Omega$ is given by
\begin{equation}\label{Laplacepdf}
p_{\mathcal{X}_\Omega^*}(\mathcal{Y}_{\Omega})
=\left(\frac{\tau}{2}\right)^{|\Omega|}\exp\left(-\frac{\|\mathcal{Y}_\Omega-\mathcal{X}_\Omega^*\|_1}{\tau}\right).
\end{equation}

Next we establish an upper error bound of the estimator
of model (\ref{model}) for the observations satisfying the noisy model (\ref{LapObN}).

\begin{theorem}\label{errLaplacNoi}
Let $\beta$ and $\lambda_i$ be defined as (\ref{betadef}) and (\ref{lambdai}), $i\in[d]$,
where $\gamma=\frac{c^2}{2\tau^2}$. Assume that the sampling
set $\Omega\sim \textup{Bern}(p)$ with $p=\frac{m}{n_1n_2\cdots n_d}$ and the joint
probability density function of $\mathcal{Y}_\Omega$ is given by
(\ref{Laplacepdf}).
Then the estimator of (\ref{model}) satisfies
\[
\begin{split}
  \frac{\mathbb{E}_{\Omega,\mathcal{Y}_{\Omega}}  \left[\|\mathcal{X}^{\lambda}-\mathcal{X}^*\|_F^2\right]}{n_1n_2\cdots n_d}
 \leq & \  \frac{11c^2(2\tau+c)^2\log(m)}{2\tau^2m} \\
 & \ + 12\left(1+\frac{2c^2}{3\tau^2}\right)(2\tau+c)^2(\beta+2)\log(n_m)\frac{r_1r_2\cdots r_d+\sum_{i=1}^d\|A_i^*\|_0}{m}.
\end{split}
\]
\end{theorem}

From Theorem \ref{errLaplacNoi}, we know that the upper error bound of the estimator of model  (\ref{model})
 is on the order of $O(\frac{r_1r_2\cdots r_d+\sum_{i=1}^d\|A_i^*\|_0}{m}\log(n_m))$.
 Similar to the case of additive Gaussian noise,
the error bound of the estimator of model  (\ref{model}) decreases as the number of observed samples increases.
Moreover, if the factor matrices of Tucker decomposition are sparser, the error bound obtained by (\ref{model})  is lower.

\begin{remark}
For the observations with additive Laplace noise,
we compared the error bound in Theorem \ref{errLaplacNoi}
with that of the matrix based  method in \cite[Corollary 5]{soni2016noisy}.
For a $d$th-order  tensor
$\mathcal{X}^*\in \mathbb{R}_+^{n_1\times\cdots\times n_d}$ with Tucker decomposition in (\ref{TucDeco}),
we unfold it into a matrix $\mathcal{X}^*_{(1)}=A_{1}^*(\mathcal{C}_{(1)}^*(\otimes_{i=d}^2A_i^*)^T)\in\mathbb{R}_+^{n_1\times (n_2\cdots n_d)}$ along the first mode, where $A_1^*$ is sparse.
The matrix based method in \cite{soni2016noisy} is then applied to $(\mathcal{X}^*_{(1)})^T\in\mathbb{R}_+^{(n_2\cdots n_d)\times n_1}$.
And the error bound of the estimator in \cite[Corollary 5]{soni2016noisy} is on
the order of $O(\frac{r_1n_2n_3\cdots n_d+\|A_1^*\|_0}{m}\log(\max\{n_1, n_2\cdots n_d\}))$.
Therefore, if $r_i$ is much smaller than $n_i, i\in[d]$,
and $d$ is not too large, the error bound of the estimator
by model (\ref{model}) is smaller than that of the matrix based method in  \cite{soni2016noisy}.
In particular, if $r_1=\cdots =r_d=r$ and $n_1=\cdots =n_d=n$,
the error bound of the estimator in Theorem \ref{errLaplacNoi} is the order of
$O(\frac{r^d+\sum_{i=1}^d\|A_i^*\|_0}{m}\log(n))$, while the error
bound of the estimator of the matrix based method in \cite{soni2016noisy} is
$O(\frac{rn^{d-1}+\|A_1^*\|_0}{m}(d-1)\log(n))$. This implies that
the error bound of the estimator in Theorem \ref{errLaplacNoi} is
smaller than that of the matrix based method in \cite{soni2016noisy}
as long as $r$ is much smaller than $n$
and $\|A_i^*\|_0$ is close to $\|A_1^*\|_0, i=2, \ldots, d$.
\end{remark}

\begin{remark}
Similar to the case for the observations with additive Gaussian noise,
we compare the error bound in Theorem \ref{errLaplacNoi} with that of the sparse nonnegative tensor factorization method
under the tensor-tensor product framework in \cite{sparse2020zhang},
which is only effective for third-order tensors.
Since the spare factors are different between the sparse NTD and sparse
nonnegative tensor factorization with tensor-tensor product in \cite{sparse2020zhang},
it is difficult to compare the error bounds of the two methods directly.
However,
if the factor  tensor is not sparse,
the error bound of the estimator in  \cite[Proposition 4.2]{sparse2020zhang} is
on the order of $O(\frac{rn_1n_3+rn_2n_3}{m}\log(\max\{n_1, n_2\}))$,
where $r$ is the tubal rank of the underlying tensor.
The error upper bound of the estimator by model  (\ref{model})
 is on the order of $O(\frac{r_1r_2r_3+\sum_{i=1}^3r_in_i}{m}\log(n_m))$,
 which is smaller than that of  \cite{sparse2020zhang} if $r$ is close to $r_i$ and $r_i$ is much smaller than $n_i$, $i=1,2,3$.
\end{remark}

\subsection{Poisson Observations}

In this subsection, we establish an upper error bound of the estimator for Poisson observations,
where the observations are modeled as
$$
\mathcal{Y}_{i_1\cdots i_d} = \textup{Poisson}(\mathcal{X}_{i_1\cdots i_d}^*), \ \ \ (i_1,\ldots, i_d)\in \Omega.
$$
Here $y = \textup{Poisson}(x)$ denotes that $y$ obeys a Poisson distribution with parameter $x>0$.
The joint probability mass function of $\mathcal{Y}_\Omega$ is given by
\begin{equation}\label{poissonpdf}
p_{\mathcal{X}_\Omega^*}(\mathcal{Y}_{\Omega})
=\prod_{(i_1,i_2,\ldots,i_d)\in\Omega}\frac{(\mathcal{X}_{i_1i_2\cdots i_d}^*)^{\mathcal{Y}_{i_1i_2\cdots i_d}}\exp({-\mathcal{X}_{i_1i_2\cdots i_d}^*})}{(\mathcal{Y}_{i_1i_2\cdots i_d})!}.
\end{equation}

Now we establish an upper error bound of the estimator of model (\ref{model}),
where the joint probability mass function of $\mathcal{Y}_\Omega$ satisfies (\ref{poissonpdf}).

\begin{theorem}\label{poissoerrb}
Suppose that $\min_{i_1,i_2,\ldots, i_d}\mathcal{X}_{i_1i_2\cdots i_d}^*\geq \varrho$
and each entry of the tensor in the candidate set $\Upsilon$ is also not smaller than $ \varrho$,
where $\varrho>0$ is a given constant.
Let $\beta$ and $\lambda_i$ be defined as (\ref{betadef}) and (\ref{lambdai}), $i\in[d]$,
where $\gamma=\frac{c^2}{\varrho}$ in (\ref{lambdai}). Assume that the sampling
set $\Omega\sim \textup{Bern}(p)$ with $p=\frac{m}{n_1n_2\cdots n_d}$ and the joint
probability mass function of $\mathcal{Y}_\Omega$ is given by
(\ref{poissonpdf}).
Then the estimator in (\ref{model}) satisfies
\[
\begin{split}
  \frac{\mathbb{E}_{\Omega,\mathcal{Y}_{\Omega}}  \left[\|\mathcal{X}^{\lambda}-\mathcal{X}^*\|_F^2\right]}{n_1n_2\cdots n_d}
 \leq & \  \frac{44c^3\log(m)}{\varrho m}  + 48c\left(1+\frac{4c^2}{3\varrho}\right)(\beta+2)\frac{r_1r_2\cdots r_d+\sum_{i=1}^d\|A_i^*\|_0}{m}\log(n_m).
\end{split}
\]
\end{theorem}

By Theorem \ref{poissoerrb},
one can obtain that the upper error bound of the estimator
of model (\ref{model}) with Poisson observations is on the order
of $O(\frac{r_1r_2\cdots r_d+\sum_{i=1}^d\|A_i^*\|_0}{m}\log(n_m))$.
Moreover, the error bound will decrease as the number of observations increases.
The lower bound of $\mathcal{X}^*$ will also influence the error bound of the estimator,
and it  will be difficult to recover $\mathcal{X}^*$ as $\varrho$ is close to zero.

\begin{remark}
In order to
compare the error bound in Theorem \ref{poissoerrb} with that of the matrix based method in \cite[Corollary 6]{soni2016noisy},
we need to unfold the underlying nonnegative tensor $\mathcal{X}^*$ $(n_1\times \cdots \times n_d)$
into a matrix, where the multi-linear structure of a tensor is ignored.
Without loss of generality, assume that $\mathcal{X}^*$ is unfolded along the first mode,
in which  the resulting matrix is
$\mathcal{X}_{(1)}^*=A_{1}^*(\mathcal{C}_{(1)}^*(\otimes_{i=d}^2A_i^*)^T)\in\mathbb{R}_+^{n_1\times (n_2\cdots n_d)}$.
Then, by applying the matrix based method in \cite{soni2016noisy} to $(\mathcal{X}_{(1)}^*)^T$,
the error bound of the resulting estimator is
on the order of $O(\frac{r_1n_2n_3\cdots n_d+\|A_1^*\|_0}{m}\log(\max\{n_1, n_2\cdots n_d\}))$,
which is larger than that of Theorem \ref{poissoerrb} if $r_i$ is much smaller than $n_i, i\in[d]$,
and $d$ is not too large.
In particular, when $r_1=\cdots =r_d=r,$ $n_1=\cdots =n_d=n$, and $\|A_1^*\|_0=\cdots=\|A_d^*\|_0$,
the error bound of the estimator in  \cite[Corollary 6]{soni2016noisy} is
on the order of $O(\frac{rn^{d-1}+\|A_1^*\|_0}{m}(d-1)\log(n))$,
and the error bound in Theorem \ref{poissoerrb}
is $O(\frac{r^d+d\|A_1^*\|_0}{m}\log(n))$,
which is smaller than that of \cite{soni2016noisy}
as long as $d\geq3$ and $r$ is much smaller than $n$.
\end{remark}

\begin{remark}
Cao et al. \cite{cao2016poisson} proposed a matrix completion method with the nuclear norm constraint
for Poisson observations and established the error bound of the estimator of their proposed model.
For higher-order tensor completion, without loss of generality,
we unfolding the underlying tensor
$\mathcal{X}^*\in\mathbb{R}_+^{n_1\times n_2\times \cdots \times n_d}$
into a matrix  along the first mode, i.e., $\mathcal{X}_{(1)}^*$.
In this case, when $m\geq (n_1+n_2\cdots n_d)\log(n_1\cdots n_d)$, the error bound of the estimator obtained by \cite[Theorem 2]{cao2016poisson} is
\begin{equation}\label{PoMaC}
\frac{\|\mathcal{X}_{mt}-\mathcal{X}^*\|_F^2}{n_1n_2\cdots n_d}\leq C\sqrt{ \frac{r_1(n_1+n_2\cdots n_d)}{m}}\log(n_1\cdots n_d)
\end{equation}
with high probability, where $\mathcal{X}_{mt}$ is the estimator obtained by \cite{cao2016poisson}
and $C>0$ is a constant.
%Therefore, if $r_i$ is much smaller than $n_i$,
In particular, when $n_1=\cdots=n_d=n$ and $r_1=\cdots =r_d$,
the upper bound in (\ref{PoMaC}) reduces to $O(d\sqrt{\frac{r_1(n+n^{d-1})}{m}}\log(n))$,
which is larger than $O(\frac{r_1^d+\sum_{i=1}^d\|A_i^*\|_0}{m}\log(n))$
if $r_1$ is much smaller than $n$ (e.g., $r_1<(\frac{n^{d-1}}{4})^{\frac{1}{2d-1}}$).
Moreover, the factor matrix $A_i$ is sparser, the error bound in Theorem \ref{poissoerrb} is lower,
while the factor matrix $A_i$ has not influence on the error bound in \cite[Theorem 2]{cao2016poisson}, $i\neq 1$.
\end{remark}

\begin{remark}
Similar to the observations with additive Laplace noise,
we can only compare the error bound in Theorem \ref{poissoerrb}
with that in \cite[Proposition 4.3]{sparse2020zhang} for third-order tensors,
which has smaller error bound than that of \cite[Theorem 3.1]{zhang2022low}
for low-rank tensor completion with Poisson observations.
More detailed comparisons about the
error bounds of the two methods can be referred  to \cite[Section IV.C]{zhang2022low}.
We do not consider the sparse case
since the factor tensor based on the algebra framework of
tensor-tensor product is not comparable with the factor matrices in Tucker decomposition.
In this case, the error bound of the estimator in  \cite[Proposition 4.3]{sparse2020zhang}
is $O(\frac{rn_1n_3+rn_2n_3}{m}\log(\max\{n_1, n_2\}))$,
where $r$ is the tubal rank of the underlying tensor.
By Theorem \ref{poissoerrb}, we obtain that
the error bound of the estimator of model  (\ref{model})
 is on the order of $O(\frac{r_1r_2r_3+\sum_{i=1}^3r_in_i}{m}\log(\max\{n_1, n_2, n_3\}))$,
 which is smaller than that of  \cite{sparse2020zhang} if $r$ is close to $r_i$ and $r_i$ is much smaller than $n_i$, $i=1,2,3$.
\end{remark}

\section{Minimax Lower Bounds}\label{lowerbou}

 In this section, we establish the minimax lower bound of an estimator for
 the observations satisfying (\ref{obserPo}).
 The accuracy of an estimator $\widetilde{\mathcal{X}}$
 for estimating the true tensor $\mathcal{X}^*$ can be measured in terms of its risk \cite{tsybakov2009},
 which is defined as
$
\frac{\mathbb{E}_{\Omega,\mathcal{Y}_{\Omega}}[\|\widetilde{\mathcal{X}}-\mathcal{X}^*\|_F^2]}{n_1n_2\cdots n_d}.
$

Now we consider a class of tensors with Tucker rank $\mathbf{r}=(r_1,\ldots, r_d)$
satisfying (\ref{TucDeco}),
where  each factor matrix in Tucker decomposition obeys $\|A_i\|_0\leq s_i, \  i\in[d]$,
and the amplitudes of each entry of the core tensor
and factor matrices are not larger than $1$ and $a_i$, respectively.
In this case, we define the following set
\begin{equation}\label{ConsSet}
 \begin{split}
 \mathfrak{L}(\mathbf{s},\mathbf{r},\mathbf{a}):=\Big\{& \mathcal{X}=\mathcal{C}	\times_1A_1\cdots\times_d A_d: \mathcal{C}\in\mathbb{R}_+^{r_1\times \cdots\times r_d},
 A_i\in\mathbb{R}_+^{n_i\times r_i}, 0\leq \mathcal{C}_{i_1\cdots  i_d}\leq 1, \|A_i\|_0\leq s_i, \\
&~~~  0\leq (A_i)_{lm}\leq a_i, (i_1,\ldots, i_d)\in[r_1]\times \cdots\times [r_d],
(l,m)\in[n_i]\times [r_i],  i\in[d] \Big\}.
\end{split}
\end{equation}

The worst-case performance of an estimator $\widetilde{\mathcal{X}}$ of $\mathcal{X}^*$
is measured  by the maximum risk on the set $\mathfrak{L}(\mathbf{s},\mathbf{r},\mathbf{a})$, which is defined as
$$
\sup_{\mathcal{X}^*\in \mathfrak{L}(\mathbf{s},\mathbf{r},\mathbf{a})}\frac{\mathbb{E}_{\Omega, \mathcal{Y}_\Omega}[\|\widetilde{\mathcal{X}}-\mathcal{X}^*\|_F^2]}{n_1n_2\cdots n_d}.
$$
The estimator, which has the smallest maximum risk among all
possible estimators, is said to achieve the minimax risk \cite{tsybakov2009}.
%which is a characteristic of the estimation problem itself.
In this case, for the sparse NTD and completion problem, the minimax risk is defined as
\begin{equation}\label{MMRLB}
\inf_{\widetilde{\mathcal{X}}}\sup_{\mathcal{X}^*\in \mathfrak{L}(\mathbf{s},\mathbf{r},\mathbf{a})}
\frac{\mathbb{E}_{\Omega, \mathcal{Y}_\Omega}[\|\widetilde{\mathcal{X}}-\mathcal{X}^*\|_F^2]}{n_1n_2\cdots n_d}.
\end{equation}

Next we will estimate the lower bound of (\ref{MMRLB}), which is stated in the following theorem.

\begin{theorem}\label{MiniMaxLowbg}
Suppose that the probability density function or probability mass function of any two entries of the noisy  observations satisfies
\begin{equation}\label{KLON}
K(\mathbb{P}_x,\mathbb{P}_y)\leq \frac{(x-y)^2}{2\mu},
\end{equation}
where $\mu>0$ depends on the noise distribution.
Assume that $r_i\leq s_i$ in the set
$\mathfrak{L}(\mathbf{s},\mathbf{r},\mathbf{a})$ in (\ref{ConsSet}), $i\in[d]$.
Then, for the joint probability density function or
probability mass function of the observations $\mathcal{Y}_\Omega$ obeying (\ref{obserPo}),
there exist two constants $\widetilde{\alpha}_m,\gamma_m>0$ such that
\[
\begin{split}
& \inf_{\widetilde{\mathcal{X}}}\sup_{\mathcal{X}^*\in \mathfrak{L}(\mathbf{s},\mathbf{r},\mathbf{a})}
\frac{\mathbb{E}_{\Omega, \mathcal{Y}_\Omega}[\|\widetilde{\mathcal{X}}-\mathcal{X}^*\|_F^2]}{n_1n_2\cdots n_d} \\
\geq \ &  \frac{\widetilde{\alpha}_m}{2^{d+5}(d+1)}\min\left\{\prod_{i=1}^d\Delta_i(s_i,n_i)a_i^2,\gamma_m^2\mu^2\left(\frac{r_1r_2\cdots r_d+\sum_{i=1}^d s_i}{m}\right) \right\},
\end{split}
\]
where
\begin{equation}\label{DSJN}
\Delta_i(s_i,n_i)=\min\left\{1,\frac{s_i}{n_i}\right\}, \ i\in[d].
\end{equation}
 \end{theorem}

For a fixed $d$, Theorem \ref{MiniMaxLowbg} shows that the lower bound
of the estimator satisfying (\ref{ConsSet}) is on the order of $O(\frac{r_1r_2\cdots r_d+\sum_{i=1}^d s_i}{m})$,
which matches to the upper bound in Theorem \ref{mainthe} up to a logarithmic factor.
This demonstrates the upper bound in  Theorem \ref{mainthe} is nearly optimal.

\begin{remark}
The proof of the minimax risk bound in Theorem \ref{MiniMaxLowbg}
utilized  the technique of the tools
for matrix completion in \cite{klopp2017robust, Sambasivan2017Minimax}, see also \cite{sparse2020zhang}.
The key issue of the proof is to construct the packing sets
for the core tensors and sparse factor matrices, respectively.
Then the Varshamov-Gilbert bound \cite[Lemma 2.9]{tsybakov2009} is used to
determine the Frobenius norm of any two tensors in the set.
Finally, one can get the minimax lower bound by the minimax
analysis techniques in \cite[Theorem 2.5]{tsybakov2009}.
\end{remark}

\begin{remark}
Based on Theorem \ref{MiniMaxLowbg},
for the explicit lower bounds of the
minimax risk with special noise models in Section \ref{upperbound},
we just need to specify $\mu$ in (\ref{KLON}) except for Poisson observations.
In particular, for the observations of sparse NTD and completion with additive Gaussian noise and additive Laplace noise,
we can set $\mu=\sigma$ and  $\mu=\tau$, respetively.
For Poisson observations, we need each entry of the
underlying tensor to be strictly positive in the construction of the packing set.
Similar results can be found in \cite{Sambasivan2017Minimax, sparse2020zhang}.
For brevity, we omit the detailed proof of the minimax risk here for sparse NTD and completion with Poisson observations.
\end{remark}

\section{ADMM Based  Algorithm}\label{Algorithm}

In this section, we design  an ADMM based algorithm \cite{boyd2011distributed} to solve problem (\ref{model}).
Notice that the feasible set $\Pi$ in (\ref{TauSet})
is discrete, which leads to the difficulty for the algorithm design.
Similar to \cite{jain2017noisy, sparse2020zhang},
we drop the discrete assumption of the core tensor and factor matrices
in order to use continuous optimization techniques,
which may be justified by choosing a very large
value of $\tau$ and by noting that continuous optimization algorithms use finite precision arithmetic
when executed on a computer.
Hence, we consider to solve the following problem:
\begin{equation}\label{ContiModel}
\begin{split}
\min_{\mathcal{X},\mathcal{C},A_i}  \ &  -\log p_{\mathcal{X}_\Omega}(\mathcal{Y}_{\Omega})+\sum_{i=1}^d\lambda_i\|A_i\|_0 \\
{\textup{s.t.}} \ & \mathcal{X}=\mathcal{C}	\times_1A_1\cdots\times_d A_d, 0\leq \mathcal{X}_{i_1i_2\cdots i_d}\leq c, (i_1,i_2,\ldots,i_d)\in[n_1]\times [n_2]\times \cdots \times [n_d], \\
 \ & 0\leq \mathcal{C}_{i_1i_2\cdots i_d}\leq 1, (i_1,i_2,\ldots,i_d)\in[r_1]\times [r_2]\times \cdots \times [r_d],  \\
 \ &  0\leq (A_i)_{ml}\leq a_i, (m,l)\in[n_i]\times [r_i], i\in[d].
\end{split}
\end{equation}

Note that the  set $\Upsilon$ in (\ref{TauSet}) and the constraint in (\ref{ContiModel}) are different.
Each entry of the core tensor $\mathcal{C}$ in $\Upsilon$  is taken from the set $\{0,\frac{1}{\tau},\frac{2}{\tau}, \ldots,\frac{\tau-1}{\tau}, 1\}$, where $\tau$ is defined in (\ref{denu}). Moreover, each entry of the factor matrix $A_i$  in $\Upsilon$  is taken from the set $\{0, \frac{a_i}{\tau},\frac{2a_i}{\tau},\ldots, \frac{(\tau-1)a_i}{\tau},a_i\}, i\in[d]$.
By relaxing the discrete assumption on the entries of the core tensor and factor matrices in $\Upsilon$, we just consider each entry of the core tensor $\mathcal{C}$ to be in $[0,1]$, i.e., $ 0\leq \mathcal{C}_{i_1i_2\cdots i_d}\leq 1, (i_1,i_2,\ldots,i_d)\in[r_1]\times [r_2]\times \cdots \times [r_d]$, and each entry of the factor matrix $A_i$ to be in $[0,a_i]$, i.e., $0\leq (A_i)_{ml}\leq a_i,  (m,l)\in[n_i]\times [r_i], i\in[d]$.
In this case, the set  $\Upsilon$ in (\ref{TauSet}) is relaxed to the constraint in (\ref{ContiModel}).

Let
\[
\begin{split}
&\Xi_1:=\left\{\mathcal{X}\in\mathbb{R}_+^{n_1\times n_2\times \cdots \times n_d}:0\leq \mathcal{X}_{i_1i_2\cdots i_d}\leq c,(i_1,i_2,\ldots,i_d)\in[n_1]\times [n_2]\times \cdots \times [n_d]\right\},  \\
& \Xi_2:=\left\{\mathcal{C}\in\mathbb{R}_+^{r_1\times r_2\times \cdots \times r_d}:0\leq \mathcal{C}_{i_1i_2\cdots i_d}\leq 1,(i_1,i_2,\ldots,i_d)\in[r_1]\times [r_2]\times \cdots \times [r_d]\right\}, \\
& \Psi_i:=\left\{A_i\in\mathbb{R}_+^{n_i\times r_i}: 0\leq (A_i)_{ml}\leq a_i,  (m,l)\in[n_i]\times [r_i]\right\}, i\in[d].
\end{split}
\]
For any set $\mathfrak{C}$, let $\delta_\mathfrak{C}(\cdot)$ denote  the indicator function over $\mathfrak{C}$, which is defined as 
$$
	\delta_\mathfrak{C}(x)
	=  \begin{cases} 0,& \textup{if} \  x \in \mathfrak{C},\\
		+\infty,&\text{otherwise.} 
	\end{cases}
$$
By using the definition of indicator function and  putting the bounded constraints with respect to $\mathcal{X},\mathcal{C},$ and $A_i$ into the objective function,
 problem (\ref{ContiModel}) can be rewritten as
\begin{equation}\label{ContiModel2}
\begin{split}
\min_{\mathcal{X},\mathcal{C},A_i}  \ &  -\log p_{\mathcal{X}_\Omega}(\mathcal{Y}_{\Omega})+\sum_{i=1}^d\lambda_i\|A_i\|_0+\delta_{\Xi_1}(\mathcal{X}) +\delta_{\Xi_2}(\mathcal{C}) + \sum_{i=1}^d \delta_{\Psi_i}(A_i)    \\
{\textup{s.t.}} \ & \mathcal{X}=\mathcal{C}	\times_1A_1\cdots\times_d A_d.
\end{split}
\end{equation}
Note that minimizing the indicator functions $\delta_{\Xi_1}(\mathcal{X}) +\delta_{\Xi_2}(\mathcal{C}) + \sum_{i=1}^d \delta_{\Psi_i}(A_i)$ is equivalent to $\mathcal{X}\in\Xi_1 ,\mathcal{C}\in \Xi_2,A_i\in \Psi_i$, $i\in[d]$.
Therefore, models (\ref{ContiModel}) and (\ref{ContiModel2}) are equivalent. 
The goal of using indicator functions in (\ref{ContiModel2}) is that there only exist equality constraints in (\ref{ContiModel2}) 
and then the ADMM can be applied to solve this model, 
where the ADMM is suitable to  solve the optimization problem with equality constraints.
Let $\mathcal{X}=\mathcal{Z},\mathcal{C}=\mathcal{B}, A_i=H_i,A_i=S_i,i\in [d]$.
Then problem (\ref{ContiModel2}) is equivalent to
\begin{equation}\label{ContiModel3}
\begin{split}
\min  \ &  -\log p_{\mathcal{X}_\Omega}(\mathcal{Y}_{\Omega})+\sum_{i=1}^d\lambda_i\|H_i\|_0+\delta_{\Xi_1}(\mathcal{Z}) +\delta_{\Xi_2}(\mathcal{B}) + \sum_{i=1}^d \delta_{\Psi_i}(S_i)    \\
{\textup{s.t.}} \ & \mathcal{X}=\mathcal{C}	\times_1A_1\cdots\times_d A_d, \mathcal{X}=\mathcal{Z},\mathcal{C}=\mathcal{B}, A_i=H_i,A_i=S_i,i\in [d].
\end{split}
\end{equation}
 Models (\ref{ContiModel2}) and (\ref{ContiModel3}) are equivalent since only the variable substitution is used in (\ref{ContiModel3}).
By introducing these variables  in (\ref{ContiModel3}), each subproblem  can be solved easily in  the framework of ADMM.
The augmented Lagrangian function associated with problem (\ref{ContiModel3}) is given by
\[
\begin{split}
& L(\mathcal{X},\mathcal{C},A_i,\mathcal{Z},\mathcal{B},H_i,S_i,\mathcal{T}_i,M_i,N_i)\\
=
 &
-\log p_{\mathcal{X}_\Omega}(\mathcal{Y}_{\Omega})+\sum_{i=1}^d\lambda_i\|H_i\|_0+\delta_{\Xi_1}(\mathcal{Z}) +\delta_{\Xi_2}(\mathcal{B}) + \sum_{i=1}^d \delta_{\Psi_i}(S_i) \\
& + \langle \mathcal{T}_1, \mathcal{X}-\mathcal{C}	\times_1A_1\cdots\times_d A_d \rangle
+\langle \mathcal{T}_2, \mathcal{X}-\mathcal{Z} \rangle +\langle \mathcal{T}_3, \mathcal{C}-\mathcal{B}\rangle \\
& + \sum_{i=1}^d \langle M_i, A_i-H_i \rangle + \sum_{i=1}^d \langle N_i, A_i-S_i \rangle \\
& +\frac{\beta_1}{2}\|\mathcal{X}-\mathcal{C}	\times_1A_1\cdots\times_d A_d\|_F^2
+\frac{\beta_2}{2}\|\mathcal{X}-\mathcal{Z}\|_F^2+\frac{\beta_3}{2}\|\mathcal{C}-\mathcal{B}\|_F^2 \\
&+\sum_{i=1}^d\left(\frac{\rho_i}{2}\|A_i-H_i\|_F^2+\frac{\alpha_i}{2}\|A_i-S_i\|_F^2\right),
\end{split}
\]
where $\mathcal{T}_1,\mathcal{T}_2,\mathcal{T}_3, M_i,N_i$ are the Lagrangian multipliers
and $\beta_1,\beta_2,\beta_3,\rho_i, \alpha_i>0$ are penalty parameters, $i\in[d]$.
The iteration template of ADMM is given as follows:
\begin{align}
             &\mathcal X^{k+1}=\mathop{\text{argmin}}_{\mathcal X} L(\mathcal{X},\mathcal{C}^k,A_i^k,\mathcal{Z}^k,\mathcal{B}^k,H_i^k,S_i^k,\mathcal{T}_i^k,M_i^k,N_i^k), \label{eq12X}
             \\
             &\mathcal C^{k+1}=\mathop{\text{argmin}}_{\mathcal C} L(\mathcal{X}^{k+1},\mathcal{C},A_i^k,\mathcal{Z}^k,\mathcal{B}^k,H_i^k,S_i^k,\mathcal{T}_i^k,M_i^k,N_i^k), \label{eq13C}
             \\
             &A_i^{k+1}=\mathop{\text{argmin}}\limits_{A_i} L(\mathcal{X}^{k+1},\mathcal{C}^{k+1},A_i,\mathcal{Z}^k,\mathcal{B}^k,H_i^k,S_i^k,\mathcal{T}_i^k,M_i^k,N_i^k), \label{eq14A}
             \\
             &\mathcal{Z}^{k+1}=\mathop{\text{argmin}}\limits_{\mathcal{Z}} L(\mathcal{X}^{k+1},\mathcal{C}^{k+1},A_i^{k+1},\mathcal{Z},
             \mathcal{B}^k,H_i^k,S_i^k,\mathcal{T}_i^k,M_i^k,N_i^k), \label{eq15Z} \\
             &\mathcal{B}^{k+1}=\mathop{\text{argmin}}\limits_{\mathcal{B}} L(\mathcal{X}^{k+1},\mathcal{C}^{k+1},A_i^{k+1},\mathcal{Z}^{k+1},\mathcal{B},H_i^k,S_i^k,\mathcal{T}_i^k,M_i^k,N_i^k), \label{eq15B} \\
             &H_i^{k+1}=\mathop{\text{argmin}}\limits_{H_i} L(\mathcal{X}^{k+1},\mathcal{C}^{k+1},A_i^{k+1},\mathcal{Z}^{k+1},\mathcal{B}^{k+1},H_i,S_i^k,\mathcal{T}_i^k,M_i^k,N_i^k), \label{eq15H} \\
             &S_i^{k+1}=\mathop{\text{argmin}}\limits_{S_i} L(\mathcal{X}^{k+1},\mathcal{C}^{k+1},A_i^{k+1},\mathcal{Z}^{k+1},\mathcal{B}^{k+1},H_i^{k+1},S_i,\mathcal{T}_i^k,M_i^k,N_i^k), \label{eq15S} \\
             & \mathcal{T}_1^{k+1} = \mathcal{T}_1^k +\beta_1(\mathcal{X}^{k+1}-\mathcal{C}^{k+1}	\times_1A_1^{k+1}\cdots\times_d A_d^{k+1}), \label{Tk1}\\
             & \mathcal{T}_2^{k+1} = \mathcal{T}_2^k +\beta_2(\mathcal{X}^{k+1}-\mathcal{Z}^{k+1}), \ \mathcal{T}_3^{k+1} = \mathcal{T}_3^k +\beta_3(\mathcal{C}^{k+1}-\mathcal{B}^{k+1}), \label{Tk23}\\
             & M_i^{k+1}=M_i^k+\rho_i(A_i^{k+1}-H_i^{k+1}), \ N_i^{k+1}= N_i^k+\alpha_i(A_i^{k+1}-S_i^{k+1}), i\in[d]. \label{MNk1}
\end{align}
The main advantage of ADMM for solving (\ref{ContiModel3}) is that each subproblem has a closed-form solution.
Next we give the explicit solutions of each subproblem in the ADMM.
Let
$$
f(\mathcal{X}):=  -\log p_{\mathcal{X}_\Omega}(\mathcal{Y}_{\Omega}).
$$
The optimal solution of (\ref{eq12X}) with respect to $\mathcal{X}$ is given by
\begin{equation}\label{Xk1}
\mathcal{X}^{k+1}=\textup{Prox}_{\frac{1}{\beta_1+\beta_2}f}\left(\frac{1}{\beta_1+\beta_2}\left(\beta_1\mathcal{C}^k	\times_1A_1^k\cdots\times_d A_d^k-\mathcal{T}_1^k+\beta_2 \mathcal{Z}^k-\mathcal{T}_2^k\right)\right).
\end{equation}

According to (\ref{VextTen}), problem (\ref{eq13C}) is equivalent to
\begin{equation}\label{COpti}
\min_\mathcal{C} \frac{\beta_1}{2}\left\|\textup{vec}\left(\mathcal{X}^{k+1}+\frac{1}{\beta_1}\mathcal{T}_1^k\right)
-(\otimes_{i=d}^1A_i^k)\textup{vec}(\mathcal{C}) \right\|_F^2+ \frac{\beta_3}{2}\left\|\textup{vec}(\mathcal{C})-\textup{vec}\left(\mathcal{B}^k-\frac{1}{\beta_3}\mathcal{T}_3^k\right)\right\|_F^2.
\end{equation}
Then the optimal solution of (\ref{COpti}) is given by
\begin{equation}\label{Ck1}
\begin{split}
\textup{vec}(\mathcal{C}^{k+1})& =\left(\beta_1(\otimes_{i=d}^1A_i^k)^T(\otimes_{i=d}^1A_i^k)+\beta_2I\right)^{-1}
\mathbf{q}^k\\
&= \left(\beta_1(\otimes_{i=d}^1(A_i^k)^TA_i^k)+\beta_2I\right)^{-1}
\mathbf{q}^k,
\end{split}
\end{equation}
where $\mathbf{q}^k:=(\otimes_{i=d}^1A_i^k)^T\textup{vec}\left(\beta_1\mathcal{X}^{k+1}+\mathcal{T}_1^k\right)
+\textup{vec}\left(\beta_3\mathcal{B}^k-\mathcal{T}_3^k\right)$ and the second equality holds by \cite[Lemma 4.2.10]{horn1991matrix}.

The problem (\ref{eq14A}) can be reformulated as
\begin{equation}\label{AProR}
\min_{A_i} \frac{\beta_1}{2}\left\|\mathcal{X}^{k+1}_{(i)}-A_iR_i^k+\frac{1}{\beta_1}(\mathcal{T}_1^k)_{(i)}\right\|_F^2
+\frac{\rho_i}{2}\left\|A_i-H_i^k+\frac{1}{\rho_i}M_i^k\right\|_F^2
\end{equation}
where
\begin{equation}\label{Rik}
R_i^k:=\mathcal{C}_{(i)}^{k+1}(A_d^k\otimes\cdots \otimes A_{i+1}^{k}\otimes A_{i-1}^{k+1}\otimes \cdots \otimes A_1^{k+1})^T.
\end{equation}
The optimal solution of (\ref{AProR}) is represented as
\begin{equation}\label{Aik1}
A_i^{k+1}=\left(\left(\beta_1\mathcal{X}^{k+1}_{(i)}+(\mathcal{T}_1^k)_{(i)}\right)(R_i^k)^T+\rho_i H_i^k-M_i^k\right)\left(\beta_1R_i^k(R_i^k)^T+\rho_i I\right)^{-1}.
\end{equation}

The optimal solution of (\ref{eq15Z}) with respect to $\mathcal{Z}$
is given by
\begin{equation}\label{Zk1}
\begin{split}
\mathcal{Z}^{k+1} & = \arg\min_{\mathcal{Z}} \delta_{\Xi_1}(\mathcal{Z})+\frac{\beta_2}{2}\left\|\mathcal{Z}-\left(\mathcal{X}^{k+1}+\frac{1}{\beta_2}\mathcal{T}_2^k\right)\right\|_F^2
=\mathbf{P}_{\Xi_1}\left(\mathcal{X}^{k+1}+\frac{1}{\beta_2}\mathcal{T}_2^k\right),
\end{split}
\end{equation}
where $\mathbf{P}_{\Xi_1}(\cdot)$ denotes the projection onto a given set $\Xi_1$.
Similarly, the optimal solution of (\ref{eq15B}) with respect to $\mathcal{B}$ is given by
\begin{equation}\label{Bk1}
\begin{split}
\mathcal{B}^{k+1}&=\arg\min_{\mathcal{B}} \delta_{\Xi_2}(\mathcal{B})
+\frac{\beta_3}{2}\left\|\mathcal{B}-\left(\mathcal{C}^{k+1}+\frac{1}{\beta_3}\mathcal{T}_3^k\right)\right\|_F^2
=\mathbf{P}_{\Xi_2}\left(\mathcal{C}^{k+1}+\frac{1}{\beta_3}\mathcal{T}_3^k\right) 
\end{split}
\end{equation}

The optimal solutions of (\ref{eq15H}) and (\ref{eq15S}) are given by
\begin{equation}\label{Hik1}
H_i^{k+1}=\arg\min_{H_i}\lambda_i\|H_i\|_0
+\frac{\rho_i}{2}\left\|H_i -\left(A_i^{k+1}+\frac{1}{\rho_i}M_i^k\right)\right\|_F^2
=\textup{Prox}_{\frac{\lambda_i}{\rho_i}\|\cdot\|_0}\left(A_i^{k+1}+\frac{1}{\rho_i}M_i^k\right)
\end{equation}
and
\begin{equation}\label{Sik1}
S_i^{k+1}= \arg\min_{S_i}\delta_{\Upsilon_i}(S_i)+\frac{\alpha_i}{2}\left\|S_i -\left(A_i^{k+1}+\frac{1}{\alpha_i}N_i^k\right)\right\|_F^2
=\mathbf{P}_{\Psi_i}\left(A_i^{k+1}+\frac{1}{\alpha_i}N_i^k\right).
\end{equation}
Although the problem in (\ref{Hik1}) is nonconvex, the minimizer of  (\ref{Hik1}) can be attained \cite[Example 6.10]{beck2017first}. 
Note that the proximal mapping of the $\ell_0$ norm of a matrix can be implemented  in a point-wise manner \cite[Theorem 6.6]{beck2017first}.
For any $\lambda>0$, the proximal mapping of $\ell_0$ norm for the one-dimension case is given by (see \cite[Example 6.10]{beck2017first})
$$
\textup{Prox}_{\lambda\|\cdot\|_0}(y)=
\begin{cases}
0,          &\text{if }  |y|<\sqrt{2\lambda}, \\
\{0,y\},    &\text{if }  |y|=\sqrt{2\lambda},\\
y,          &\text{if }  |y|>\sqrt{2\lambda}.
\end{cases}
$$
Therefore, by  \cite[Theorem 6.6]{beck2017first}, the minimizer of  (\ref{Hik1}) is given by
\begin{equation}\label{Hpoiwis}
(H_i^{k+1})_{jt}=
\begin{cases}
	0,          &\text{if }  |(A_i^{k+1}+\frac{1}{\rho_i}M_i^k)_{jt}|<\sqrt{\frac{2\lambda_i}{\rho_i}}, \\
	\{0,(A_i^{k+1}+\frac{1}{\rho_i}M_i^k)_{jt}\},    &\text{if }  |(A_i^{k+1}+\frac{1}{\rho_i}M_i^k)_{jt}|=\sqrt{\frac{2\lambda_i}{\rho_i}},\\
	(A_i^{k+1}+\frac{1}{\rho_i}M_i^k)_{jt},          &\text{if }  |(A_i^{k+1}+\frac{1}{\rho_i}M_i^k)_{jt}|>\sqrt{\frac{2\lambda_i}{\rho_i}}.
\end{cases}
\end{equation}

Now the ADMM for solving (\ref{ContiModel3}) is stated
  in Algorithm \ref{algorithm2}.

\begin{algorithm}[htbp]
    \caption{Alternating Direction Method of
    Multipliers for Solving Problem (\ref{ContiModel3}).} \label{algorithm2}
    \begin{algorithmic}[1]
        \State Initialization: $\mathcal{C}^0, A_i^0, \mathcal{Z}^0,\mathcal{B}^0,
        \mathcal{T}_1^0,\mathcal{T}_2^0,\mathcal{T}_3^0, M_i^0, N_i^0, i\in[d]$. Given parameters $\lambda_i,\beta_1,\beta_2,\rho_i,\alpha_i>0,i\in[d]$.
        \Repeat
  		\State \textbf{Step 1.} Compute ${\mathcal X}^{k+1}$ by (\ref{Xk1}).
  		\State \textbf{Step 2.} Update ${\mathcal C}^{k+1}$ by (\ref{Ck1}).
  		\State \textbf{Step 3.} Compute $A_i^{k+1}$ via (\ref{Aik1}), $i\in[d]$.
  		\State \textbf{Step 4.} Compute $\mathcal{Z}^{k+1}$   by (\ref{Zk1}).
        \State \textbf{Step 5.} Update $\mathcal{B}^{k+1}$   according to (\ref{Bk1}).
        \State \textbf{Step 6.} Compute $H_i^{k+1}$ and $S_i^{k+1}$  by (\ref{Hpoiwis}) and (\ref{Sik1}), respectively, $i\in[d]$.
        \State \textbf{Step 7.} Update $\mathcal{T}_1^{k+1}$, $\mathcal{T}_2^{k+1}$,
        $\mathcal{T}_3^{k+1}$, $M_i^{k+1}$, $N_i^{k+1}$   by (\ref{Tk1}), (\ref{Tk23}) and  (\ref{MNk1}), respectively.
        \Until A stopping condition is satisfied.
	\end{algorithmic}
\end{algorithm}

\begin{remark}
For the projections in (\ref{Zk1}),  (\ref{Bk1}), and (\ref{Sik1}), the component of each entry   keeps the same if it is in the set, otherwise projects to the nearest boundary.
For instance, for the projection in  (\ref{Zk1}), 
\[
\begin{split}
(\mathcal{Z}^{k+1})_{i_1i_2\cdots i_d}
& =\left(\mathbf{P}_{\Xi_1}\left(\mathcal{X}^{k+1}+\frac{1}{\beta_2}\mathcal{T}_2^k\right)\right)_{i_1i_2\cdots i_d} \\
&=\begin{cases}
0,          &\text{if }  (\mathcal{X}^{k+1}+\frac{1}{\beta_2}\mathcal{T}_2^k)_{i_1i_2\cdots i_d}<0, \\
(\mathcal{X}^{k+1}+\frac{1}{\beta_2}\mathcal{T}_2^k)_{i_1i_2\cdots i_d},    &\text{if }  0\leq (\mathcal{X}^{k+1}+\frac{1}{\beta_2}\mathcal{T}_2^k)_{i_1i_2\cdots i_d}\leq c,\\
c,          &\text{if }  (\mathcal{X}^{k+1}+\frac{1}{\beta_2}\mathcal{T}_2^k)_{i_1i_2\cdots i_d}>c,
\end{cases}
\end{split}
\]
where $(i_1,i_2,\ldots,i_d)\in[n_1]\times [n_2]\times \cdots \times [n_d]$.
\end{remark}

In the implementation of Algorithm \ref{algorithm2},
one needs to compute the proximal mapping of $\frac{1}{\beta_1+\beta_2}f$ at $\mathcal{H}$,
where
\begin{equation}\label{HKSno}
\mathcal{H}:=\frac{1}{\beta_1+\beta_2}\left(\beta_1\mathcal{C}^k	\times_1A_1^k\cdots\times_d A_d^k-\mathcal{T}_1^k+\beta_2 \mathcal{Z}^k-\mathcal{T}_2^k\right).
\end{equation}
In particular, for the special noise  observation models including
additive Gaussian noise, additive Laplace noise, and Poisson observations,
the proximal mapping in (\ref{Xk1}) is given in detail in \cite{sparse2020zhang}, which is summarized as follows.
\begin{itemize}
\item Additive Gaussian noise: 
$
\mathcal{X}^{k+1} = \frac{1}{1+\sigma^2(\beta_1+\beta_2)}
\mathcal{P}_{\Omega}\left(\mathcal{Y}+\sigma^2(\beta_1+\beta_2)\mathcal{H}\right)+\mathcal{P}_{\overline{\Omega}}(\mathcal{H}),
$
where  $\overline{\Omega}$ is the complementary set of $\Omega$
 and $\mathcal{P}_\Omega$ is the projection operator onto the index $\Omega$ such that
$$
\left(\mathcal{P}_\Omega\left(\mathcal{A}\right)\right)_{i_1i_2\cdots i_d}
=  \begin{cases} \mathcal{A}_{i_1i_2\cdots i_d},& \textup{if} \  (i_1,i_2,\cdots, i_d) \in \Omega,\\
0,&\text{otherwise.} \end{cases}
$$
\item Additive Laplace noise: 
$
\mathcal{X}^{k+1} =\mathcal{P}_{\Omega}(\mathcal{Y}+\textup{sign}(\mathcal{H}-\mathcal{Y})\circ \max\{|\mathcal{H}-\mathcal{Y}|-\frac{1}{\tau(\beta_1+\beta_2)},0\}) +\mathcal{P}_{\overline{\Omega}}(\mathcal{H}),
$
where $\textup{sign}(\cdot)$ and $\circ $ represents the signum function and the Hadamard  point-wise product, respectively.
\item Poisson observations:   
$$
\mathcal{X}^{k+1} = \frac{1}{2(\beta_1+\beta_2)}\mathcal{P}_{\Omega}((\beta_1+\beta_2)\mathcal{H}-\mathbf{1}+\sqrt{((\beta_1+\beta_2)\mathcal{H}-\mathbf{1})^2
+4(\beta_1+\beta_2)\mathcal{Y}})+\mathcal{P}_{\overline{\Omega}}(\mathcal{H}),
$$
where $\mathbf{1}$ represents an $n_1\times n_2\times \cdots \times n_d$ tensor with all entries being $1$,
the square root and square are performed in the point-wise manner.
\end{itemize}

The  computational complexity of ADMM  at each iteration for solving (\ref{ContiModel3}) is given as follows.
If the observations are chosen as additive Gaussian noise, additive Laplace noise, or Poisson observations,
the computation cost of $\mathcal{X}^{k+1}$ is on the order of $O(\sum_{j=1}^d(\prod_{i=1}^jn_i)(\prod_{i=j}^dr_i))$.
The computational cost of $\mathcal{C}^{k+1}$ is
$O(\sum_{i=1}^d r_i^2n_i+\prod_{i=1}^dr_i^3+\sum_{j=1}^d(\prod_{i=1}^jr_i)(\prod_{i=j}^dn_i))$.
In the implementation of $R_i^k$ in  (\ref{Rik}), we do not compute the Kronecker product directly.
Let
$
\mathcal{K}_i:= \mathcal{C}^{k+1}	\times_1A_1^{k+1}\cdots\times_{i-1} A_{i-1}^{k+1} \times_{i+1} A_{i+1}^k\times \cdots\times_d A_d^k.
$
Then $(\mathcal{K}_{i})_{(i)}=R_i^k$.
Note that the computational cost of $\mathcal{K}_i$ is \cite[Appendix B]{xu2015alternating}
\begin{equation}\label{ComKi}
O\left(\sum_{j=1}^{i-1}\left(\prod_{t=1}^jn_t\right)\left(\prod_{t=j}^d r_t\right)+r_i\left(\prod_{t=1}^{i-1}n_t\right)\sum_{j=i+1}^{d}\left(\prod_{t={i+1}}^jn_t\right)\left(\prod_{t=j}^d r_t\right)\right).
\end{equation}For any $i\in[d]$,
the computational cost of each factor matrix $A_i$ is the sum of (\ref{ComKi}) and
$
O(r_in_1\cdots n_d+r_i^3+ n_ir_i^2).
$
The computational costs of $\mathcal{Z}^{k+1}$ and $\mathcal{B}^{k+1}$
are $O(\prod_{i=1}^d n_i)$ and $O(\prod_{i=1}^d r_i)$, respectively.
The computational costs of $H_i^{k+1}$ and $S_i^{k+1}$ are both $O(n_ir_i)$.
The computational costs of the
multipliers $\mathcal{T}_1,\mathcal{T}_2,\mathcal{T}_3,M_i,N_i$ are
$O(\sum_{j=1}^d(\prod_{i=1}^jn_i)(\prod_{i=j}^dr_i))$.
Note that $r_i\leq n_i,i\in[d]$,
and the quantities of (\ref{ComKi}) and $O(\sum_{j=1}^d(\prod_{i=1}^jr_i)(\prod_{i=j}^dn_i))$ are similar, where the detailed discussions can be found  in \cite[Appendix B]{xu2015alternating}.
Therefore, the computational cost of ADMM in Algorithm \ref{algorithm2} is
$$
O\left(d\sum_{j=1}^d\left(\prod_{i=1}^jn_i\right)\left(\prod_{i=j}^dr_i\right)
+\sum_{j=1}^d\left(\prod_{i=1}^jr_i\right)\left(\prod_{i=j}^dn_i\right)+\sum_{i=1}^d r_i^2n_i+\prod_{i=1}^dr_i^3+\left(\prod_{i=1}^dn_i\right)\sum_{i=1}^dr_i\right).
$$

\begin{remark}
 Notice that model (\ref{ContiModel3}) is nonconvex since the $\ell_0$ norm  and the constraint
	$\mathcal{X}=\mathcal{C}	\times_1A_1\cdots\times_d A_d$ are both nonconvex.
The convergence of ADMM cannot be guaranteed in general.
Although great efforts have been made about the convergence of ADMM for solving nonconvex problems,
the constraints and objective in (\ref{ContiModel3})  are both nonconvex,
where the objective is not Lipschitz continuous for general loss functions mentioned in Section \ref{upperbound}.
Therefore, the existing literature about the convergence
of ADMM for nonconvex problems (e.g., see \cite{boct2020proximal, hong2016convergence, wang2019global})
 cannot be applied to our model directly.
\end{remark}

\section{Numerical Experiments}\label{NumerExper}

In this section, we evaluate the effectiveness of the proposed sparse NTD and completion model (SNTDC)
using both synthetic data and real image data.
We compare SNTDC with
a matrix based method in \cite{soni2016noisy} (denoted by Matrix)
and sparse nonnegative tensor factorization and completion with tensor-tensor product (SNTFTP) \cite{sparse2020zhang}. All experiments are performed in MATLAB 2020a on a computer with  Intel Xeon W-2133 and  32 GB of RAM.
 In particular, the two comparison methods are introduced in detail below. 
\begin{itemize}
\item A low-rank matrix factorization method with bounded constraints for a  general class of noisy matrix
completion tasks is proposed in \cite{soni2016noisy},
where the underlying matrix is approximated by the product of two matrices and one matrix is sparse.
This approach can be reduced to sparse nonnegative matrix factorization and completion with specially bounded constraints on the matrices, where the corresponding
model is given by
\[
\begin{split}
	\min_{X,A,B} & \  -\log p_{{X}_\Omega}({Y}_{\Omega})+\lambda_1\|{B}\|_0 \\
	\textup{s.t.} &  \  {X}= {A} {B}, 0\leq \mathcal{X}_{ij}\leq c_1,  0\leq  {A}_{ij}\leq 1, 0\leq  {B}_{ij}\leq b_1,
\end{split}
\]
Here $p_{{X}_\Omega}({Y}_{\Omega})$ is the same as that in  (\ref{obserPo}) for the case of  second-order tensors, and $\lambda_1,c_1,b_1>0$ are given constants.
Then the error bounds for the estimator of the above model is established, which is applied to the special observations including additive Gaussian noise, additive Laplace noise,
Poisson-distributed observations, and highly quantized  observations.
\item The SNTFTP in \cite{sparse2020zhang} is a sparse nonnegative tensor factorization method with partial and noisy observations for third-order tensors based on tensor-tensor product, where one factor tensor is sparse and all entries of the two factor tensors are nonnegative and bounded. 
More specifically, the SNTFTP model is given by
\[
\begin{split}
\min_{\mathcal{X},\mathcal{A},\mathcal{B}} & \  -\log p_{\mathcal{X}_\Omega}(\mathcal{Y}_{\Omega})+\lambda_2\|\mathcal{B}\|_0 \\
\textup{s.t.} &  \ \mathcal{X}=\mathcal{A}\diamond\mathcal{B}, 0\leq \mathcal{X}_{ijk}\leq c_2,  0\leq \mathcal{A}_{ijk}\leq 1, 0\leq \mathcal{B}_{ijk}\leq b_2,
\end{split}
\]
where $\mathcal{A}\diamond\mathcal{B}$ denotes the tensor-tensor product of the two third-order tensors $\mathcal{A}, \mathcal{B}$ defined in  \cite{Kilmer2011Factorization}, and $\lambda_2,c_2,b_2>0$ are given constants.
Besides, the error bound  of the estimator of
the SNTFTP  model is established under general noise
observations.
And the minimax lower bound is shown to be matched with the
established upper bound  up to a logarithmic factor of the sizes
of the underlying tensor. 
\end{itemize}

Algorithm \ref{algorithm2} will be stopped if the maximum number of iterations $300$ is reached or the following condition is satisfied
$$
\frac{\|\mathcal{X}^{k+1}-\mathcal{X}^k\|_F}{\|\mathcal{X}^{k}\|_F}\leq 10^{-4},
$$
where $\mathcal{X}^k$ is the $k$th iteration of Algorithm \ref{algorithm2}.
For the initialization of Algorithm \ref{algorithm2}, we use a sequentially truncated high order singular value decomposition \cite{nick2012new}
on $\mathcal{P}_{\Omega}(\mathcal{Y})$ to get the initial values $\mathcal{C}^0,A_i^0, i\in[d]$, which can contain some information of $\mathcal{X}$.
For $\mathcal{B}^0$, we choose it the same as $\mathcal{C}^0$.  
We set $\mathcal{Z}^0=
\mathcal{T}_1^0=\mathcal{T}_2^0=\mathcal{T}_3^0=\mathcal{P}_{\Omega}(\mathcal{Y})$ and $ M_i^0=N_i^0=A_i^0, i\in[d]$.
For the regularization parameters $\lambda_i$, we set $\lambda_i$ to
the same for any $i\in[d]$ and choose them from the set $\{5,10,50\}$ to get the best recovery performance.
Moreover, $\beta_i$ are set the same for different $i$ and chosen from the set $\{50,70,100,250,350,450,550\}$.
For $\alpha_i$ and $\rho_i$, we set them the same in each case for any $i\in[d]$,
and choose them from the set $\{0.1,0.01,0.001\}$ to get the best recovery performance.

\subsection{Synthetic Data}

In this subsection, we test the synthetic data to demonstrate the effectiveness of the proposed method.
The synthetic tensor is generated as follows:
The underlying tensor has the Tucker decomposition
$
\mathcal{X}=\mathcal{C}	\times_1 A_1\times_2 A_2\cdots \times_d A_d,
$
where the nonnegative core tensor $\mathcal{C}$ is generated by MATLAB command $\textup{rand}([n_1,\ldots, n_d])$,
and the sparse nonnegative factor matrices $A_i$ are generated by $a_i\cdot\textup{sptenrand}([n_i,r_i],\upsilon_i)$, $i\in[d]$.
Here $\upsilon_i$ is the sparse ratio of $A_i$, i.e., $\upsilon_i=\frac{\|A_i\|_0}{n_ir_i}, i\in[d]$.
We set $c = 2\|\mathcal{X}\|_{\infty}$.
For these synthetic data, we test 10 trials in each case
and the average result of these trials  is the last result.
The relative error is employed to measure the accuracy of different approaches, which is defined as ${\|\widetilde{\mathcal{X}}_0-\mathcal{X}^*\|_F}/{\|\mathcal{X}^*\|_F}$. 
Here $\mathcal{X}_0$ and $\mathcal{X}^*$ represent the recovered tensor and ground-truth tensor, respectively.

\begin{figure}[!t]
	\centering
	\subfigure[Gaussian noise]{
    		\includegraphics[width=5.3cm,height=4cm]{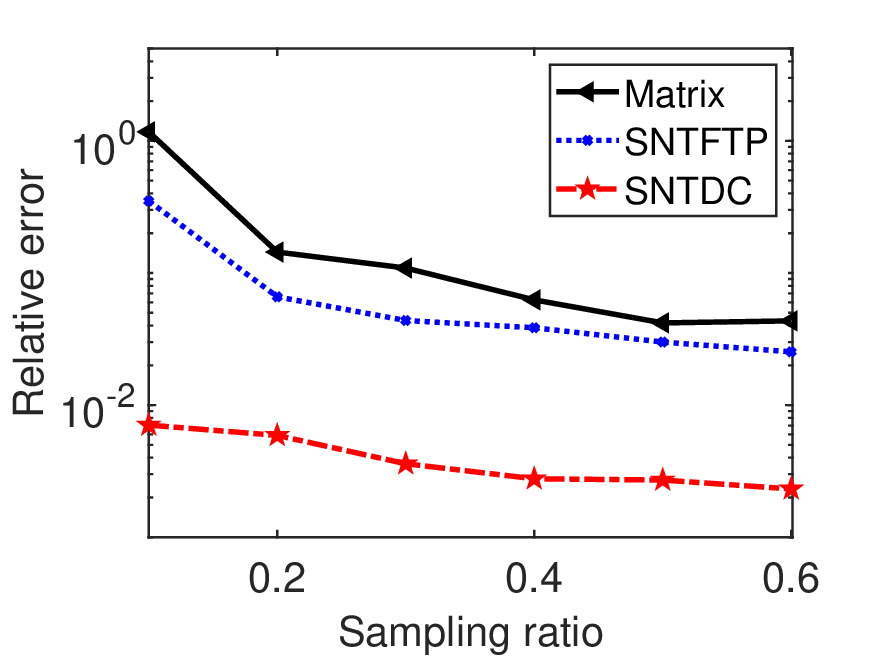}
		%\caption{fig1}
	}\hspace{-5mm}	
%		\quad
	\subfigure[Laplace noise]{
    		\includegraphics[width=5.3cm,height=4cm]{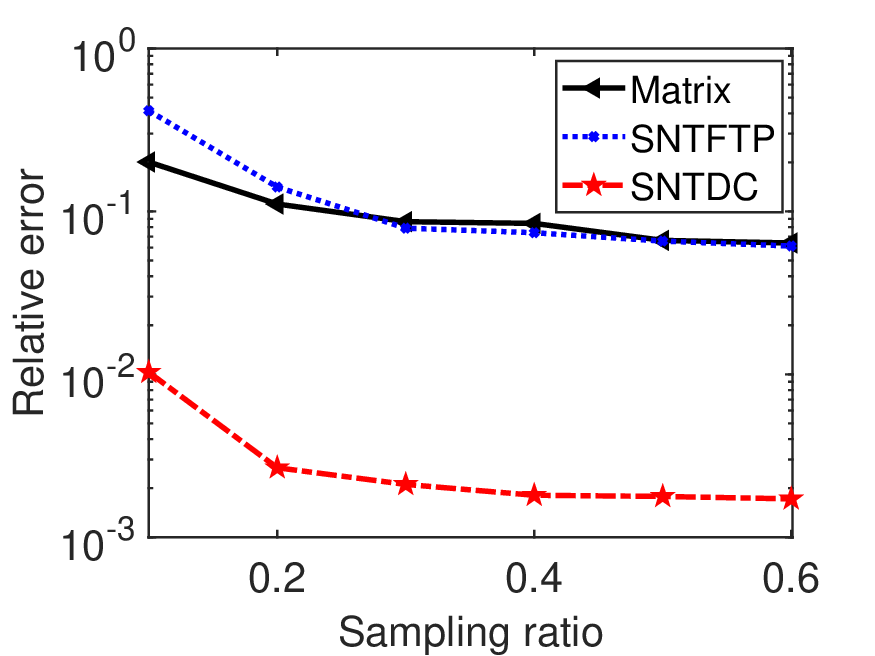}
		%\caption{fig1}
	}\hspace{-5mm}	
%		\quad
	\subfigure[Poisson observations]{
    		\includegraphics[width=5.3cm,height=4cm]{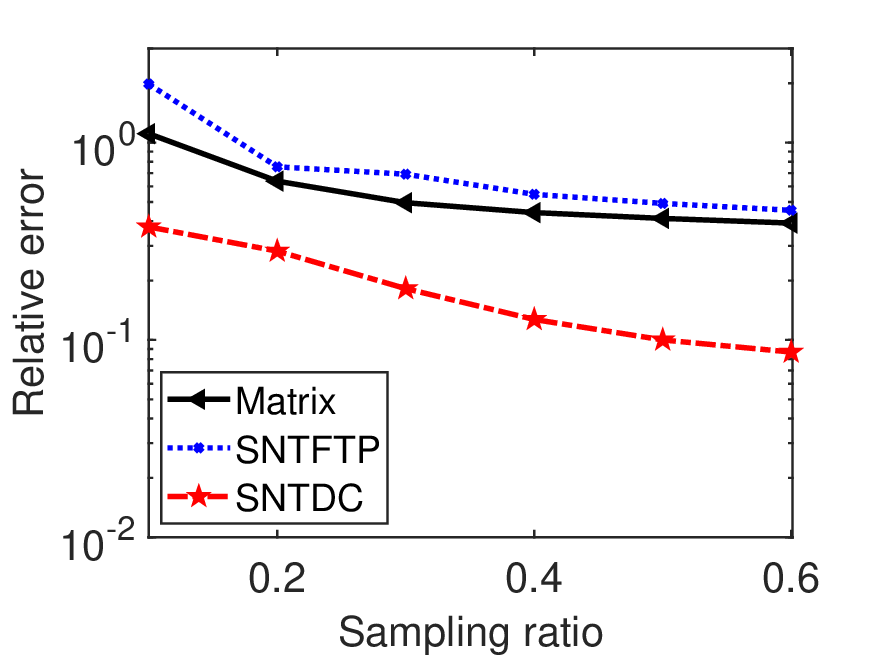}
		%\caption{fig1}
	}\hspace{-5mm}	
	\caption{\small Relative error versus sampling ratio for synthetic tensors
with size $100\times 100\times 100$  and Tucker rank $(5,5,5)$.}
	\label{Gaussin3order}	
\end{figure}

We first test the third-order tensors with $n_1=n_2=n_3=100$, whose Tucker rank is $(5,5,5)$.
For the observations with additive Gaussian noise and additive Laplace noise,
we set $\sigma^2=0.01$ and $\tau=0.01$, respectively.
In Figure \ref{Gaussin3order}, we show the relative error versus sampling ratios
of different methods for the observations with additive Gaussian noise, additive Laplace noise,
and Poisson observations, respectively,
where $\upsilon_i=0.3, i=1,2,3$.
It can be observed that the relative errors of the matrix based method, SNTFTP, and SNTDC decrease
as the sampling ratio increases.
Moreover, the relative errors obtained by SNTDC are smaller than those
obtained by the matrix based method and SNTFTP for the three kinds of observations.

\begin{figure}[!t]
	\centering
	\subfigure[Gaussian noise]{
    		\includegraphics[width=5.3cm,height=4cm]{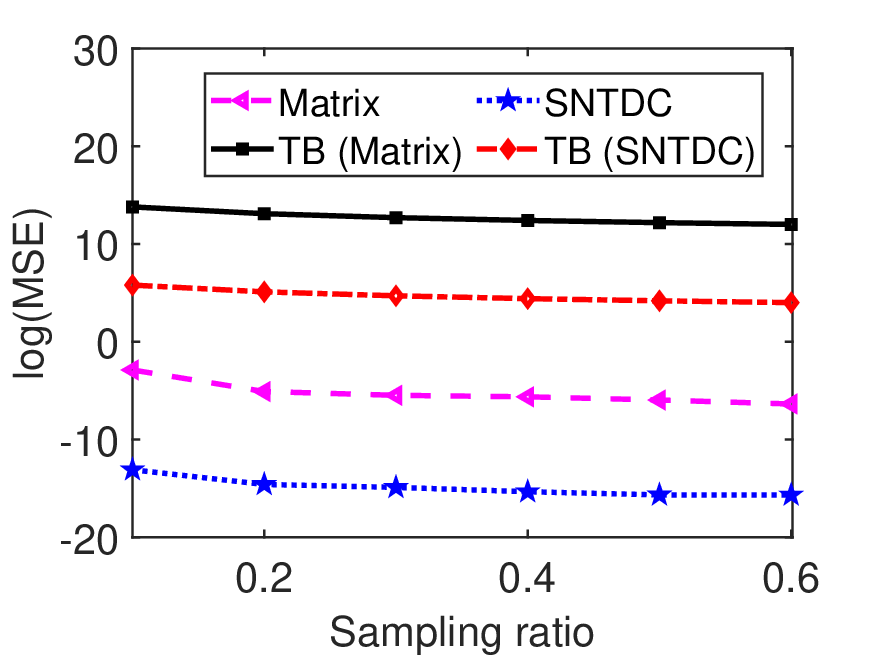}
		%\caption{fig1}
	}\hspace{-5mm}	
%		\quad
	\subfigure[Laplace noise]{
    		\includegraphics[width=5.3cm,height=4cm]{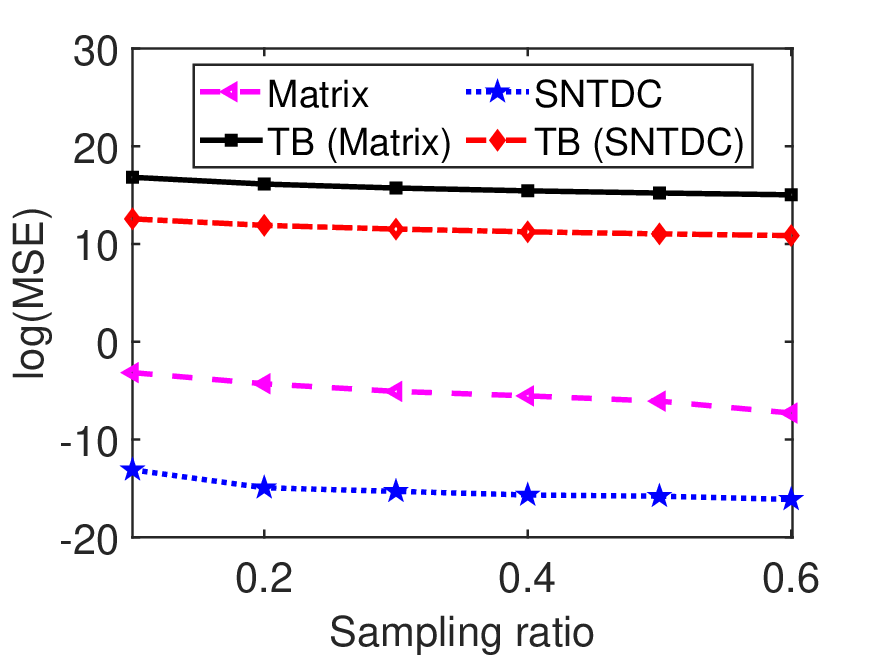}
		%\caption{fig1}
	}\hspace{-5mm}	
%		\quad
	\subfigure[Poisson observations]{
    		\includegraphics[width=5.3cm,height=4cm]{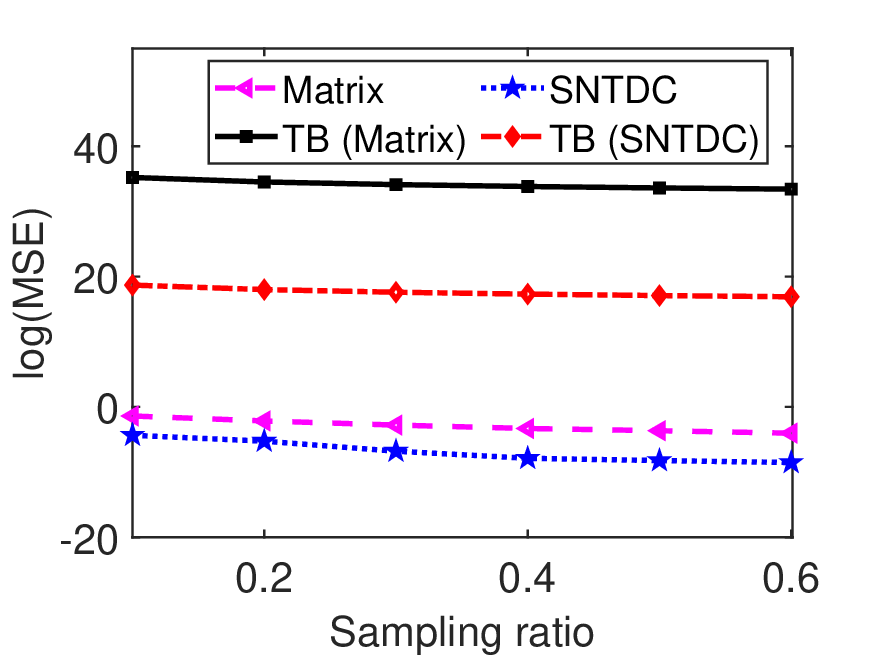}
		%\caption{fig1}
	}\hspace{-5mm}	
	\caption{\small Plots of MSE on a logarithmic scale versus sampling ratio for synthetic tensors
with size $50\times 50\times 50\times 50$  and Tucker rank $(5,5,5,5)$.}\label{GaussLP4order}	
\end{figure}

 Now we test a fourth-order tensor
$50\times 50\times 50 \times 50$ with Tucker rank $(5,5,5,5)$,
where $\sigma^2=0.01$ for additive Gaussian noise,  $\tau=0.01$ for additive Laplace noise,
and $\upsilon_i=0.3, i=1,2,3$.
Since the SNTFTP is only effective for third-order tensors, we do not compare with it for fourth-order tensors in this experiment.
The mean squared error (MSE) is used to measure the recovery accuracy of different methods, which is defined as $\textup{MSE}=\frac{\|\widetilde{\mathcal{X}}_1-\mathcal{X}^*\|_F^2}{n_1\cdots n_d}$. Here $\widetilde{\mathcal{X}}_1$ and $\mathcal{X}^*$ denote the recovered tensor and the ground-truth tensor, respectively.
Besides, we also compare the theoretical  bounds of the matrix based method in \cite{soni2016noisy} and SNTDC for different noise models, which are denoted by TB (Matrix) and TB (SNTDC), respectively.
In Figure \ref{GaussLP4order}, we show
the MSE on a logarithmic scale versus sampling ratio of the theoretical and testing results for the models with additive Gaussian noise, additive Laplace noise, and Poisson observations, respectively, where the sampling ratio is set from $0.1$ to $0.6$ with step size $0.1$.
Naturally, in each case, the errors decrease as the sampling ratios increase.
Furthermore, the theoretical bounds of SNTDC are lower than those of the matrix based method in \cite{soni2016noisy} for the three different noise models.
In the testing experiments, 
the errors of SNTDC are
much smaller than those of the matrix based method for the observations with different noises.
However,  the theoretical bounds of SNTDC are larger than those of the experimental cases, which are similar for the matrix based method.
The main reason for this phenomenon is that  some inequalities are used  to estimate the error bounds for both the matrix based method and SNTDC, which are not tight.

\subsection{Image Data}

In this subsection, we test the Swimmer
dataset\footnote{\footnotesize \url{https://stodden.net/Papers.html}} ($32\times 32\times 256$) \cite{NIPS2003_1843e35d}
and  the Columbia Object Image Library
(COIL-100)\footnote{\url{https://www.cs.columbia.edu/CAVE/software/softlib/coil-100.php}}
for sparse nonnegative Tucker decomposition and completion.
For the Swimmer dataset,  the size of each image is $32\times 32$
and there are 256 swimmer images.
Similar to \cite{xu2015alternating}, the Tucker rank of
the Swimmer dataset is set to $(24,20,20)$ in Algorithm \ref{algorithm2}.
For the COIL-100,
which contains 100 objects,
we resize each image to $64\times 64\times 3$ and choose  first 50 images for the fifty-sixth  object,
In this case, the size of the resulting tensor is $64\times 64\times 3\times 50$,
whose Tucker rank is set to $(15,15,3,10)$ in Algorithm \ref{algorithm2}.

\begin{figure}[!t]
	\centering
	\subfigure[Gaussian noise]{
    		\includegraphics[width=5.3cm,height=4cm]{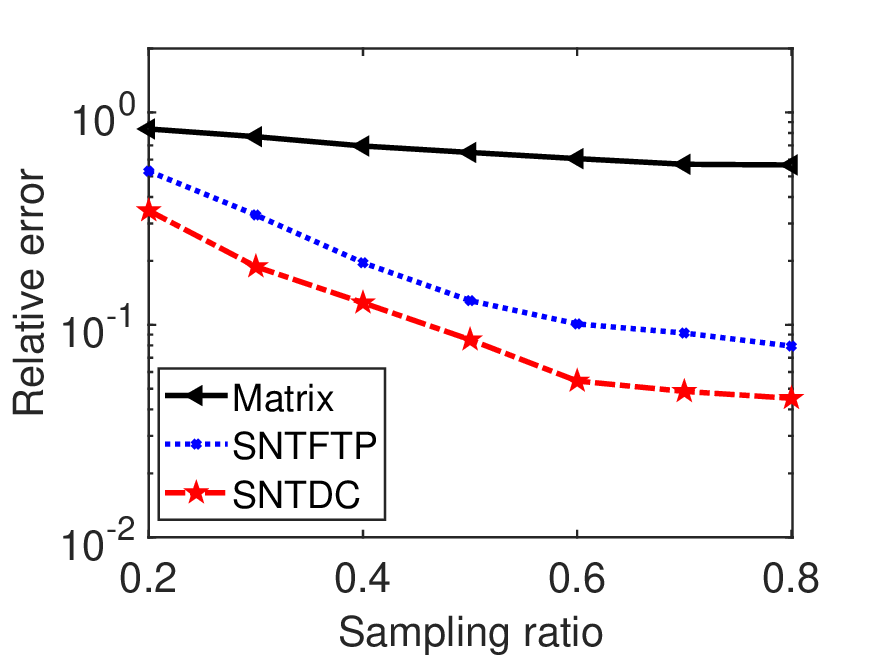}
		%\caption{fig1}
	}\hspace{-5mm}	
%		\quad
	\subfigure[Laplace noise]{
    		\includegraphics[width=5.3cm,height=4cm]{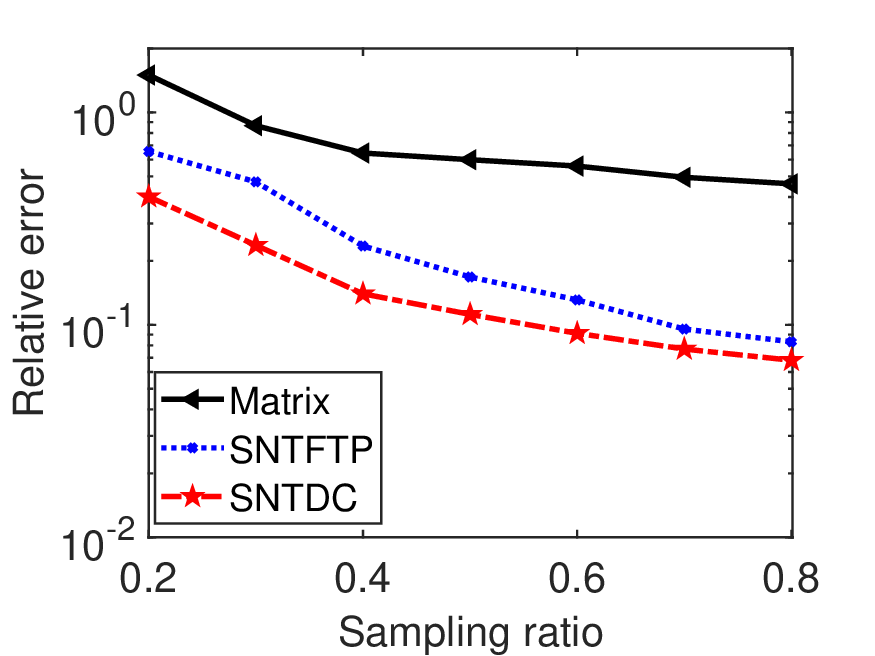}
		%\caption{fig1}
	}\hspace{-5mm}	
%		\quad
	\subfigure[Poisson observations]{
    		\includegraphics[width=5.3cm,height=4cm]{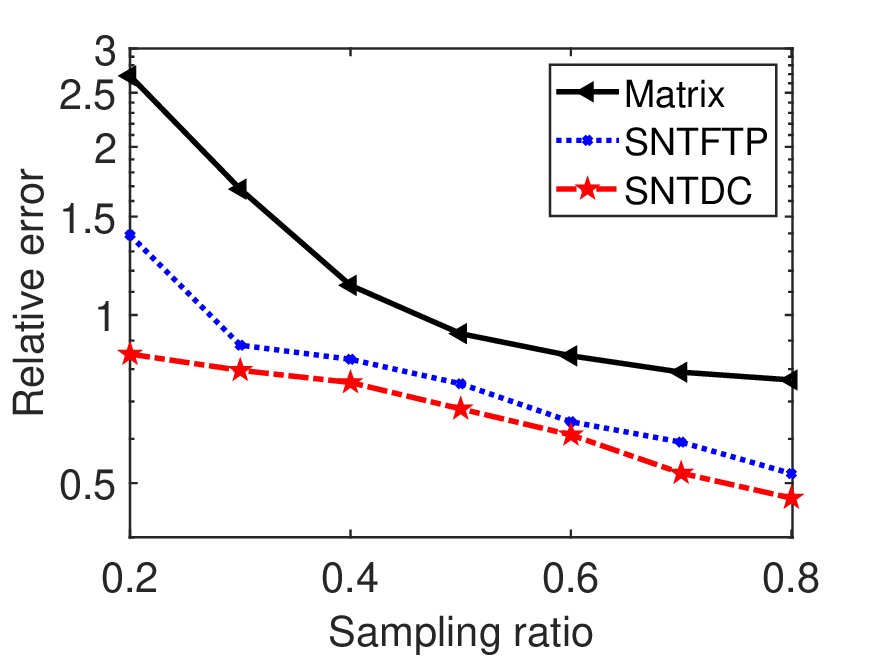}
		%\caption{fig1}
	}\hspace{-5mm}	
	\caption{\small Relative error versus sampling ratio for the Swimmer dataset.}\label{GaussLPReal}	
\end{figure}

In Figure \ref{GaussLPReal}, we show the relative error versus sampling ratio
of different methods for the observations with  additive Gaussian noise,
additive Laplace noise, and Poisson observations, respectively,
where $\sigma^2=0.01$ for additive Gaussian noise and $\tau=0.01$ for additive Laplace noise.
It can be seen from this figure that the relative errors of the SNTDC are smaller than
those of the matrix based method and SNTFTP for different sampling ratios.
Moreover, the performance of SNTFTP is better than that of the matrix based method in terms of  relative errors for the three kinds of observations.
And the relative errors of the matrix based method, SNTFTP, and SNTDC decrease
as the number of samples increases for the observations with additive Gaussian noise,
additive Laplace noise, and Poisson observations.

\begin{figure}[!t]
	\centering
	\subfigure[Gaussian noise]{
    		\includegraphics[width=5.3cm,height=4cm]{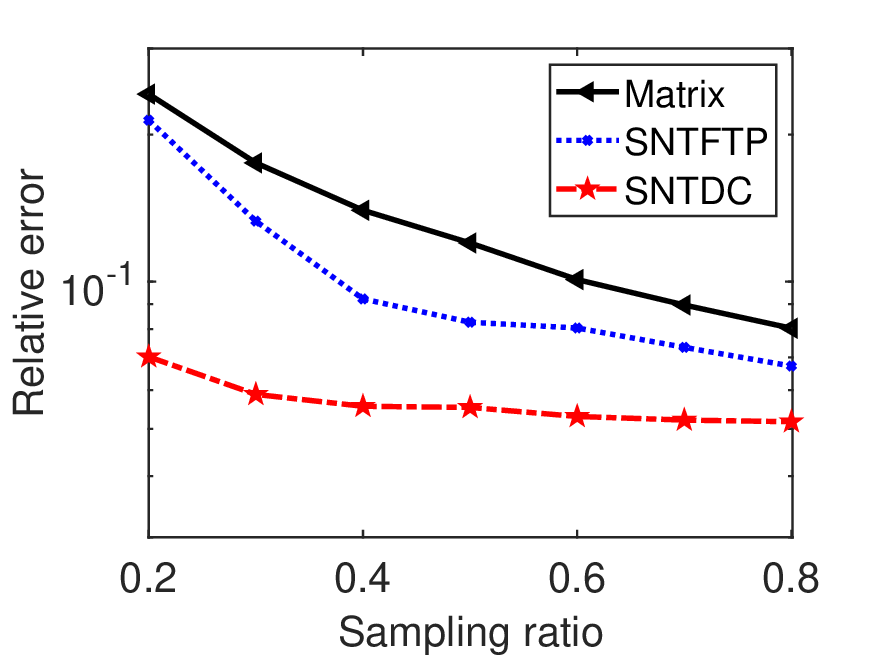}
		%\caption{fig1}
	}\hspace{-5mm}	
%		\quad
	\subfigure[Laplace noise]{
    		\includegraphics[width=5.3cm,height=4cm]{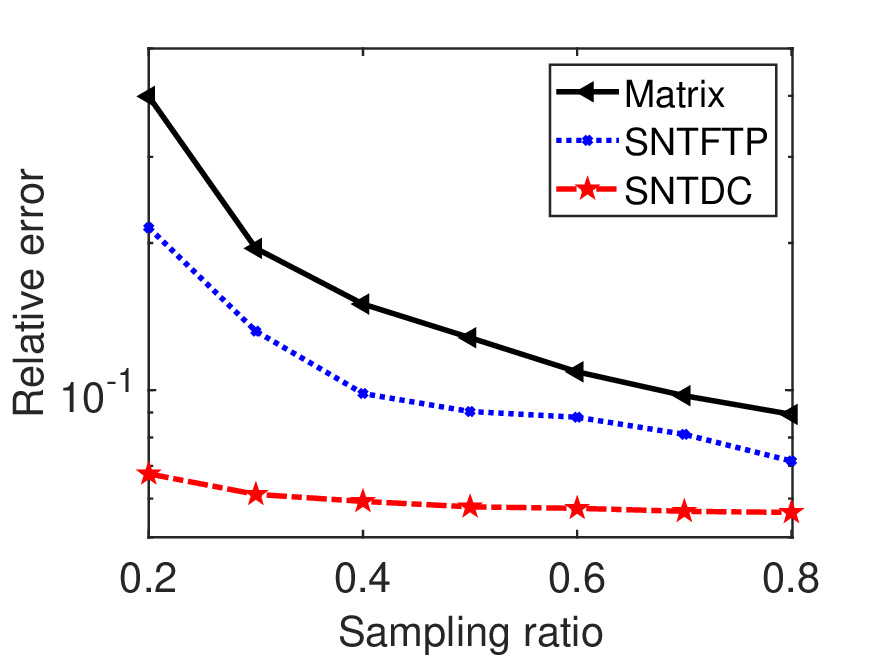}
		%\caption{fig1}
	}\hspace{-5mm}	
%		\quad
	\subfigure[Poisson observations]{
    		\includegraphics[width=5.3cm,height=4cm]{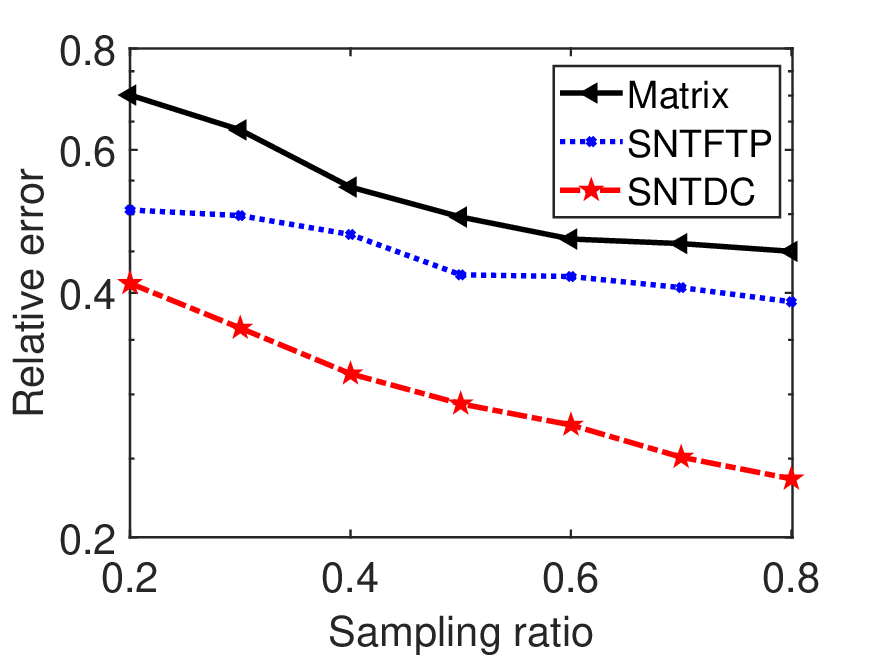}
		%\caption{fig1}
	}\hspace{-5mm}	
	\caption{\small Relative error versus sampling ratio for the COIL-100 dataset.}\label{GaussLPReal4thorder}	
\end{figure}

Since the SNTFTP is only effective for third-order tensors,
we stack all images of the COIL-100 based on the color channels
into a third-order tensors along the frontal slices for the SNTFTP,
where the size of the resulting tensor is $ 64\times 64\times 150$.
In Figure \ref{GaussLPReal4thorder}, we display the
relative error versus sampling ratio of different methods for the COIL-100 dataset,
where $\sigma^2=0.05$ for additive Gaussian noise and $\tau=0.05$ for additive Laplace noise.
We can see that the relative errors obtained by the matrix based method,
SNTFTP, and SNTDC decrease as the sampling ratio increases for different noise observations.
And the relative errors obtained by SNTDC are
smaller than those obtained by the matrix based method and SNTFTP.
Moreover, the SNTFTP performs better than the matrix based method
in terms of relative error for the observations with
additive Gaussian noise, additive Laplace noise, and Poisson observations.

\section{Conclusions and Future Work}\label{Conclu}

In this paper, we have investigated the problem of sparse
NTD and completion with a general class of noise observations,
where the underlying tensor is decomposed into a core tensor
and several factor matrices  with nonnegativity and sparsity constraints.
Moreover, the loss function is derived by the maximum likelihood estimation of the noisy observations.
And the error bound of the estimator of the SNTDC model is established under a class of noise distributions.
Then the error bounds are specified to some widely used noise models
including additive Gaussian noise, additive Laplace noise, and Poisson observations.
Besides, the minimax lower bound of the observed model is derived, which matches to the upper bound up to a logarithmic factor.
Numerical experiments on synthetic data and real-world data sets demonstrate the superior performance of the proposed
sparse NTD and completion model compared with other methods.

In the experiments, we need to set the Tucker rank of the underlying tensor in advance,
which is unknown in general for real-world data sets.
An interesting direction for future work on this problem
is  to update the Tucker rank adaptively.
Besides, in the sparse NTD and completion model,  the $\ell_0$ norm is employed to
characterise the sparsity of the factor matrices.
However, the $\ell_0$ norm is nonconvex and it is challenge to get the unique and global optimal solution.
Therefore, it is also of great interest to replace the $\ell_0$ norm
by the $\ell_1$ norm for the factor matrices and then analyze the error bound of the corresponding model.

The convergence of Algorithm \ref{algorithm2} cannot be guaranteed in general
since our model in (\ref{ContiModel3}) is nonconvex and multi-block.
Future work is devoted to establishing  the convergence of ADMM for our nonconvex multi-block  model.
Moreover, due to effectiveness and explanatory information of
 orthogonal constraints on the
factor matrices for nonnegative Tucker decomposition (c.f. \cite{pan2021orthogonal}),
it will also be of great interest to extend the error bound
of our model to that of orthogonal nonnegative Tucker decomposition under general loss functions.

\section*{Appendix A. Proof of Theorem \ref{mainthe}}

At the beginning, the following lemma is first recorded, which plays a vital role for the proof of Theorem \ref{mainthe}.

\begin{lemma}\label{leapp}
Let
$\Upsilon$ be a countable collection of candidate
reconstructions $\mathcal{X}$ of $\mathcal{X}^*$ and its penalty $\emph{pen}(\mathcal{X})\geq 1$
satisfying $\sum_{\mathcal{X}\in\Upsilon}2^{-\textup{pen}(\mathcal{X})}\leq 1$.
For any integer $2^d\leq m\leq n_1n_2\cdots n_d$, let $\Omega\sim  \textup{Bern}(p)$.
Moreover, the joint probability density/mass function of the corresponding observations follows
$p_{\mathcal{X}_{\Omega}^*}(\mathcal{Y}_{\Omega})=\prod_{(i,j,k)\in\Omega}p_{\mathcal{X}_{ijk}^*}(\mathcal{Y}_{ijk})$,
which are assumed to be conditionally independent on the given $\Omega$.
Let $\gamma$ be a constant satisfying
\begin{equation}\label{kappar}
\gamma\geq \max_{\mathcal{X}\in\Upsilon}\max_{i_1,i_2,\ldots, i_d} K\left(p_{\mathcal{X}_{i_1i_2\cdots i_d}^*}(\mathcal{Y}_{i_1i_2\cdots i_d})||p_{\mathcal{X}_{i_1i_2\cdots i_d}}(\mathcal{Y}_{i_1i_2\cdots i_d})\right).
\end{equation}
Consider the following complexity penalized maximum likelihood estimator
\begin{equation}\label{mxial}
{\mathcal{X}}^\mu\in\arg\min_{\mathcal{X}\in\Upsilon}\left\{-\log p_{\mathcal{X}_\Omega}(\mathcal{Y}_{\Omega})+\mu\cdot\emph{pen}(\mathcal{X})\right\}.
\end{equation}
Then for any
$\mu\geq2\left(1+\frac{2\gamma}{3} \right) \log(2)$,
one has
\[
\begin{split}
&  \ \frac{\mathbb{E}_{\Omega,\mathcal{Y}_{\Omega}}\left[-2\log H(p_{{\mathcal{X}}^\mu}(\mathcal{Y}),p_{\mathcal{X}^*}(\mathcal{Y}))\right]}{n_1n_2\cdots n_d} \\
\leq & \  3\cdot\min_{\mathcal{X}\in\Upsilon}\left\lbrace \frac{K(p_{\mathcal{X}^*}(\mathcal{Y})|| p_{\mathcal{X}}(\mathcal{Y})}{n_1n_2\cdots n_d}+\left( \mu+\frac{4\gamma\log(2)}{3}\right)\frac{\emph{pen}(\mathcal{X})}{m} \right\rbrace +\frac{8\gamma\log(m)}{m},
\end{split}
\]
where the expectation is taken with respect to the joint distribution of $\Omega$ and $\mathcal{Y}_{\Omega}$.
\end{lemma}

The proof of Lemma \ref{leapp} can be obtained
easily based on the matrix case \cite[Lemma 8]{soni2016noisy} and \cite{li1999estimation},
see also for the cases of  CP decomposition in \cite[Lemma 1]{jain2017noisy}
and tensor factorization via tensor-tensor product in \cite[Lemma 1.1]{sparse2020zhang}.
The three steps of proof in \cite[Lemma 8]{soni2016noisy} are giving the ``good'' sample set characteristics,
a conditional error guarantee and some simple conditioning arguments,
which are in point-wise manners in fact for the KL divergence,
negative logarithmic Hellinger affinity, and maximum likelihood estimation.
As a consequence, we can extend them to the tensor case with Tucker decomposition easily.
Here we omit the details.

The proof of Theorem \ref{mainthe} follows the line of the proof of \cite[Theorem 1]{soni2016noisy},
see also the proofs of \cite[Theorem 3]{raginsky2010compressed} and \cite[Theorem 4.1]{sparse2020zhang}.
The main technique of this proof is the well-known Kraft-McMillan inequality \cite{Brockway1956Two, kraft1949device}.
The penalty of the underlying tensor $\mathcal{X}$ is constructed by the codes of the core tensors and sparse factor matrices,
where the underlying tensor has the Tucker decomposition form
 with the core tensor being nonnegative and the factor matrices being nonnegative and sparse.
Next we return to the proof of Theorem \ref{mainthe}.

Based on the result in Lemma \ref{leapp}, one should define the penalties
 $\textup{pen}(\mathcal{X})\geq 1$ in the candidate reconstructions $\mathcal{X}$ of $\mathcal{X}^*$
such that the penalties $\mathcal{X}$ in $\Upsilon$ satisfy
\begin{equation}\label{KMMI}
\sum_{\mathcal{X}\in\Upsilon}2^{-\textup{pen}(\mathcal{X})} \leq 1.
\end{equation}
The condition (\ref{KMMI}) is the
well-known Kraft-McMillan inequality for coding elements of $\Upsilon$ with an alphabet of size $2$, which is satisfied automatically
if we choose the penalties to be code lengths for the unique decodable binary code for the elements $\mathcal{X}\in \Upsilon$ \cite{kraft1949device, Brockway1956Two}, see also \cite[Section 5]{cover2006elements}.
This also gives the constructions of the penalties.

Let
$
\Upsilon_1:=\{\mathcal{X}=\mathcal{C}	\times_1A_1\cdots\times_d A_d:
\ \mathcal{C}\in\mathfrak{C}, \ A_i\in\mathfrak{B}_i\},
$
where $\mathfrak{C}$ and $\mathfrak{B}_i, i\in[d]$ are the same as those constructed  in Section \ref{ProMod}.
Note that $\Upsilon_1\subseteq\Upsilon$ by the construction of $\Upsilon_1$.
Now we consider the discretized core tensor $\mathcal{C}\in\mathfrak{C}$ and sparse factor matrices $A_i\in\mathfrak{B}_i, i\in[d]$.
\begin{itemize}
\item[(1)] We encode the amplitude of each element
of $\mathcal{C}$ using $\log_2(\tau)$ bits, where $\tau$ is defined as (\ref{denu}).
Then a total of $r_1r_2\cdots r_d\log_2({\tau})$ bits are used to encode the core tensor $\mathcal{C}$.

\item[(2)] Let $\zeta_i:=2^{\lceil\log_2(r_in_i)\rceil}, i\in[d]$.
We encode each nonzero element of $A_i$ using $\log_2(\zeta_i)$ to denote the location and $\log_2(\tau)$ bits for its amplitude.
In this case,
a total of $\|A_i\|_0(\log_2(\tau)+\log_2(\zeta_i))$ bits
is used to encode the sparse factor matrix $A_i$.
\item[(3)] Finally, we can assign each $\mathcal{X}\in\Upsilon_1$ whose code length satisfies
$$
\textup{pen}(\mathcal{X})=r_1r_2\cdots r_d\log_2({\tau})+\sum_{i=1}^d\|A_i\|_0(\log_2(\tau)+\log_2(\zeta_i)).
$$
By the constructions, we know that such codes are uniquely decodable, which implies that $\sum_{\mathcal{X}\in\Upsilon_1}2^{-\textup{pen}(\mathcal{X})} \leq 1$ \cite{kraft1949device, Brockway1956Two}.
Since $\Upsilon\subseteq\Upsilon_1$, one has
$$
\sum_{\mathcal{X}\in\Upsilon}2^{-\textup{pen}(\mathcal{X})} \leq \sum_{\mathcal{X}\in\Upsilon_1}2^{-\textup{pen}(\mathcal{X})} \leq 1.
$$
\end{itemize}
Consequently, by Lemma \ref{leapp}, we get that the estimator $\mathcal{X}^\mu$ in the following
\begin{equation}
\begin{split}
\mathcal{X}^\mu & \in\arg\min_{\mathcal{X}\in\Upsilon}\left\{-\log p_{\mathcal{X}_\Omega}(\mathcal{Y}_{\Omega})+\mu \cdot\textup{pen}(\mathcal{X})\right\} \\
& = \arg\min_{\mathcal{X}\in\Upsilon}\left\{-\log p_{\mathcal{X}_\Omega}(\mathcal{Y}_{\Omega})+\mu \sum_{i=1}^d\|A_i\|_0(\log_2(\tau)+\log_2(\zeta_i))\right\},
\end{split}
\end{equation}
satisfies
\[
\begin{split}
& \frac{\mathbb{E}_{\Omega,\mathcal{Y}_{\Omega}}\left[-2\log H(p_{{\mathcal{X}}^\mu},p_{\mathcal{X}^*})\right]}{n_1n_2\cdots n_d}
\leq  \frac{8\gamma\log(m)}{m}+ 3\cdot\min_{\mathcal{X}\in\Upsilon}\Biggl\{\frac{K(p_{\mathcal{X}^*}(\mathcal{Y})|| p_{\mathcal{X}}(\mathcal{Y})}{n_1n_2\cdots n_d}
+\\
& ~~~~~~~~~~~~~~~~~~~~~~~~~~~~~~~~\left( \mu+\frac{4\gamma\log(2)}{3}\right)\frac{r_1r_2\cdots r_d\log_2(\tau)+\sum_{i=1}^d\|A_i\|_0(\log_2(\tau)+\log_2(\zeta_i))}{m} \Biggl\},
\end{split}
\]
where $\mu\geq2\left(1+\frac{2\gamma}{3} \right) \log(2)$ and $\gamma$ satisfies (\ref{kappar}).

Let $\lambda_i=\mu(\log_2(\tau)+\log_2(\zeta_i)), i\in[d]$.
Note that $r_i\leq n_i$ and
\begin{equation}\label{LGHDT}
\log_2(\tau)+\log_2(\zeta_i)\leq 2(\beta+2)\frac{\log(n_m)}{\log(2)}.
\end{equation}
Therefore, for any
$
\lambda_i\geq 4(\beta+2)(1+\frac{2\gamma}{3})\log(n_m), i\in[d],
$
the estimator of
\begin{equation}
\begin{split}
{\mathcal{X}}^{\lambda} \in
 \arg\min_{\mathcal{X}\in\Upsilon}\left\{-\log p_{\mathcal{X}_\Omega}(\mathcal{Y}_{\Omega})+\sum_{i=1}^d\lambda_i\|A_i\|_0\right\}
\end{split}
\end{equation}
satisfies
\[
\begin{split}
& \  \frac{\mathbb{E}_{\Omega,\mathcal{Y}_{\Omega}}\left[-2\log H(p_{{\mathcal{X}}^{\lambda}}(\mathcal{Y}),p_{\mathcal{X}^*}(\mathcal{Y}))\right]}{n_1n_2\cdots n_d}
\leq \frac{8\gamma\log(m)}{m}+ 3\cdot\min_{\mathcal{X}\in\Upsilon}\Biggl\{ \frac{K(p_{\mathcal{X}^*}(\mathcal{Y})|| p_{\mathcal{X}}(\mathcal{Y}))}{n_1n_2\cdots n_d}\\
& \ ~~~~~~~~~~~~~~~~~~~~~~~~~~~~~~~~~~ +\left(\max_i\{\lambda_i\}+\frac{8\gamma(\beta+2)\log(n_m)}{3}\right)\frac{r_1r_2\cdots r_d+\sum_{i=1}^d\|A_i\|_0}{m} \Biggl\},
\end{split}
\]
where the inequality holds by (\ref{LGHDT}).
This completes the proof. \qed

\section*{Appendix B. Proof of Theorem \ref{errspeGaNo}}

First, we establish an upper bound of the tensor infinity norm
between the underlying tensor $\mathcal{X}^*$ and its closest
surrogate in $\Upsilon$, where $\mathcal{X}^*=\mathcal{C}^*	\times_1 A_1^*\times_2 A_2^*\cdots \times_d A_d^*$,
and $\mathcal{C}^*,A_i^*\in[d]$ are defined as (\ref{Constecf}).
The following estimate will be useful in the sequel.

\begin{lemma}\label{InfNor}
Let $\mathcal{X}^*_s=\mathcal{C}^s	\times_1A_1^s\cdots\times_d A_d^s$, where the entries of  $\mathcal{C}^s$ are the closest discretized surrogates of the entries of $\mathcal{C}^*$ in $\Upsilon$,
and the entries of $A_i^s$ are the closest discretized
surrogates of the entries of $A_i^*$ in $\Upsilon$, $i\in[d]$. Then
 it holds that
$$
\|\mathcal{X}^*_s-\mathcal{X}^*\|_\infty \leq \frac{2^{d+1}-1}{\tau-1}\prod_{i=1}^d (a_ir_i).
$$
where $\tau$ is defined as  (\ref{denu}).
\end{lemma}
{\bf Proof}.
Let $\mathcal{C}^s=\mathcal{C}^*+\triangle\mathcal{C}$
and $A_i^s=A_i^*+\triangle A_i, i\in[d]$.
By the definition of $\mathcal{C}^*$ and $A_i^*,i\in[d]$ in (\ref{Constecf}),
we get that  $\|\triangle\mathcal{C}\|_\infty\leq \frac{1}{\tau-1}$
and $\|\triangle A_i\|_\infty\leq \frac{a_i}{\tau-1}$.
It follows from  (\ref{VextTen}) that
\begin{equation}\label{XXSV}
\begin{split}
 \mathcal{X}^*_s-\mathcal{X}^*
=& \ (\otimes_{i=d}^1(A_i^*+\triangle A_i))\textup{vec}(\mathcal{C}^*+\triangle\mathcal{C})-(\otimes_{i=d}^1A_i^*)\textup{vec}(\mathcal{C}^*)\\
=& \ (\otimes_{i=d}^1(A_i^*+\triangle A_i))\textup{vec}(\mathcal{C}^*)+(\otimes_{i=d}^1(A_i^*+\triangle A_i))(\triangle\mathcal{C})-(\otimes_{i=d}^1A_i^*)\textup{vec}(\mathcal{C}^*).
\end{split}
\end{equation}
Furthermore, we have
\[
\begin{split}
 \ & \otimes_{i=d}^1(A_i^*+\triangle A_i)\\
= \ &  A_d^*\otimes\cdots\otimes A_1^* +   \triangle A_d\otimes A_{d-1}^*  \otimes\cdots\otimes A_1^* + \cdots
+   A_d^* \otimes\cdots\otimes A_2^*\otimes\triangle A_1 \\
\ & + \triangle A_d\otimes \triangle A_{d-1} \otimes A_{d-3}^* \otimes\cdots\otimes A_1^* + \cdots +
 A_d^* \otimes\cdots\otimes A_3^*\otimes \triangle A_2\otimes\triangle A_1 \\
\ & +\cdots \\
\ & + \triangle A_d \otimes\cdots\otimes \triangle A_{2}  \otimes A_1^* + \cdots
+   A_d^* \otimes\triangle A_{d-1}\otimes\cdots\otimes \triangle A_1 \\
\ &  + \triangle A_d\otimes\cdots\otimes\triangle A_1.
\end{split}
\]
Since $0\leq (A_i^*)_{lm}\leq a_i$ and $0\leq \mathcal{C}^*_{i_1i_2\cdots i_d}\leq 1$, $\forall (l,m)\in[n_i]\times [r_i], (i_1,i_2,\ldots,i_d)\in[r_1]\times [r_2]\times \cdots \times [r_d], i\in[d]$, we have that
\begin{equation}\label{CDETE}
\begin{split}
&\|(A_d^*\otimes\cdots\otimes A_1^*)\textup{vec}(\mathcal{C}^*)\|_\infty \leq \prod_{i=1}^d (a_ir_i), \\
& \| (\triangle A_d\otimes A_{d-1}^*  \otimes\cdots\otimes A_1^* + \cdots
+   A_d^* \otimes\cdots\otimes A_2^*\otimes\triangle A_1)\textup{vec}(\mathcal{C}^*)\|_\infty\leq \frac{C_d^1}{\tau-1}\prod_{i=1}^d (a_ir_i), \\
& ~~~~~~~~~~~~~~~~~~~~~ \vdots \\
& \|(A_d^* \otimes\triangle A_{d-1}\otimes\cdots\otimes \triangle A_1)\textup{vec}(\mathcal{C}^*)\|_\infty \leq \frac{C_d^{d-1}}{(\tau-1)^{d-1}}\prod_{i=1}^d (a_ir_i), \\
& \|(\triangle A_d\otimes\cdots\otimes\triangle A_1)\textup{vec}(\mathcal{C}^*)\|_\infty \leq \frac{1}{(\tau-1)^{d}}\prod_{i=1}^d (a_ir_i),
\end{split}
\end{equation}
which yields that
\begin{equation}\label{1TDEB}
\|(\otimes_{i=d}^1(A_i^*+\triangle A_i))\textup{vec}(\mathcal{C}^*)-(\otimes_{i=d}^1A_i^*)\textup{vec}(\mathcal{C}^*)\|_\infty \leq \left(1+\frac{1}{\tau-1}\right)^d\prod_{i=1}^d (a_ir_i)-\prod_{i=1}^d (a_ir_i).
\end{equation}
Similarly, by  $\|\triangle\mathcal{C}\|_\infty\leq \frac{1}{\tau-1}$, we can easily get that
\begin{equation}\label{2EBC}
\|(\otimes_{i=d}^1(A_i^*+\triangle A_i))(\triangle\mathcal{C})\|_\infty\leq \frac{1}{\tau-1}\left(1+\frac{1}{\tau-1}\right)^d\prod_{i=1}^d (a_ir_i).
\end{equation}
Note that
\begin{equation}\label{InQDH}
\begin{split}
\left(1+\frac{1}{\tau-1}\right)^{d+1}- 1 & = C_{d+1}^1\frac{1}{\tau-1}+C_{d+1}^2\frac{1}{(\tau-1)^2}+\cdots + C_{d+1}^{d+1}\frac{1}{(\tau-1)^{d+1}} \\
& =\frac{1}{\tau-1}\left(C_{d+1}^1+C_{d+1}^2\frac{1}{\tau-1}+\cdots + C_{d+1}^{d+1}\frac{1}{(\tau-1)^{d}} \right)\\
& \leq \frac{1}{\tau-1}\left(C_{d+1}^1+C_{d+1}^2+\cdots + C_{d+1}^{d+1}\right) =\frac{2^{d+1}-1}{\tau-1},
\end{split}
\end{equation}
where the inequality holds by $\tau\geq 2$.
Therefore, combining (\ref{XXSV}), (\ref{CDETE}), (\ref{1TDEB}) and (\ref{2EBC}), we obtain that
\begin{equation}
\begin{split}
 \|\mathcal{X}^*_s-\mathcal{X}^*\|_\infty
 & \leq \left(1+\frac{1}{\tau-1}\right)^d\prod_{i=1}^d (a_ir_i)-\prod_{i=1}^d (a_ir_i) + \frac{1}{\tau-1}\left(1+\frac{1}{\tau-1}\right)^d\prod_{i=1}^d (a_ir_i)\\
% &\leq r_1r_2\cdots r_d\left(\left(1+\frac{1}{\tau-1}\right)\left(a_1+\frac{a_1}{\tau-1}\right)\cdots\left(a_d+\frac{a_d}{\tau-1}\right)-a_1a_2\cdots a_d\right)\\
&=\left(\left(1+\frac{1}{\tau-1}\right)^{d+1}- 1\right) \prod_{i=1}^d (a_ir_i) \\
&\leq \frac{2^{d+1}-1 }{\tau-1}\prod_{i=1}^d (a_ir_i),
\end{split}
\end{equation}
where the last inequality follows form  (\ref{InQDH}).
%\[
%\begin{split}
%\left(1+\frac{1}{\tau-1}\right)^{d+1}- 1 & = C_{d+1}^1\frac{1}{\tau-1}+C_{d+1}^2\frac{1}{(\tau-1)^2}+\cdots + C_{d+1}^{d+1}\frac{1}{(\tau-1)^{d+1}} \\
%& =\frac{1}{\tau-1}\left(C_{d+1}^1+C_{d+1}^2\frac{1}{\tau-1}+\cdots + C_{d+1}^{d+1}\frac{1}{(\tau-1)^{d}} \right)\\
%& \leq \frac{1}{\tau-1}\left(C_{d+1}^1+C_{d+1}^2+\cdots + C_{d+1}^{d+1}\right) =\frac{2^{d+1}-1}{\tau-1}.
%\end{split}
%\]
This furnishes the desired statement. \qed

Now we return to the proof of Theorem \ref{errspeGaNo}.
Recall that the negative logarithmic Hellinger affinity
and KL divergence for additive Gaussian noise are (see, e.g., \cite{soni2016noisy, sparse2020zhang})
$$
-\log(H(p_{\mathcal{X}_{i_1i_2\cdots i_d}}
(\mathcal{Y}_{i_1i_2\cdots i_d}),p_{(\mathcal{X}^*)_{i_1i_2\cdots i_d}}(\mathcal{Y}_{i_1i_2\cdots i_d})))
=\frac{(\mathcal{X}_{i_1i_2\cdots i_d}-(\mathcal{X}^*)_{i_1i_2\cdots i_d})^2}{8\sigma^2}
$$
and
$$
K\left(p_{\mathcal{X}_{i_1i_2\cdots i_d}^*}(\mathcal{Y}_{i_1i_2\cdots i_d})||p_{\mathcal{X}_{i_1i_2\cdots i_d}}(\mathcal{Y}_{i_1i_2\cdots i_d})\right)=\frac{(\mathcal{X}_{i_1i_2\cdots i_d}-\mathcal{X}_{i_1i_2\cdots i_d}^*)^2}{2\sigma^2}.
$$
Therefore, we obtain that
\begin{equation}\label{AKLFG}
-2\log H(p_{{\mathcal{X}}^{\lambda}}(\mathcal{Y}),p_{\mathcal{X}^*}(\mathcal{Y}))=\frac{\|\mathcal{X}^\lambda-\mathcal{X}^*\|_F^2}{4\sigma^2}.
\end{equation}
Since $\mathcal{X}\in \Upsilon$ and $0\leq \mathcal{X}_{i_1i_2\cdots i_d}^*\leq \frac{c}{2}$,
we can take $\gamma=\frac{c^2}{2\sigma^2}$ in (\ref{kappar}).
Therefore, it follows from Theorem \ref{mainthe} that
\begin{equation}\label{ChaTheGa}
\begin{split}
& \  \frac{\mathbb{E}_{\Omega,\mathcal{Y}_{\Omega}}\left[\|\mathcal{X}^\lambda-\mathcal{X}^*\|_F^2\right]}{n_1n_2\cdots n_d}
\leq \frac{16c^2\log(m)}{m}+ 12\sigma^2\cdot\min_{\mathcal{X}\in\Upsilon}\Biggl\{ \frac{K(p_{\mathcal{X}^*}(\mathcal{Y})|| p_{\mathcal{X}}(\mathcal{Y}))}{n_1n_2\cdots n_d}\\
& \ ~~~~~~~~~~~~~~~~~~~~~~~~~~~~~~~~~~ +\left(\max_i\{\lambda_i\}+\frac{4c^2(\beta+2)\log(n_m)}{3\sigma^2}\right)\frac{r_1r_2\cdots r_d+\sum_{i=1}^d\|A_i\|_0}{m} \Biggl\}.
\end{split}
\end{equation}
By the definition of $\beta$ in (\ref{betadef}), we know that
\begin{equation}\label{taulowb}
\begin{split}
\tau=2^{\lceil\log_2(n_m)^\beta\rceil}& \geq 2^{\beta\log_2(n_m)}  \geq 2^{ \log_2(n_m)+ \log_2\left(\frac{(2^{d+1}-1)\sqrt{d}r_1r_2\cdots r_d a_1a_2\cdots a_d}{c \sqrt{n_m}}+1\right)} \\
& \geq \frac{(2^{d+1}-1)\sqrt{dn_m}r_1r_2\cdots r_d a_1a_2\cdots a_d}{c}+1.
\end{split}
\end{equation}
%where the last inequality holds by $n_m\geq 2$.
For any $\mathcal{X}^s$ constructed in Lemma \ref{InfNor},
  we have
$$
\|\mathcal{X}^s-\mathcal{X}^*\|_\infty\leq \frac{c}{\sqrt{dn_m}}\leq \frac{c}{2},
$$
where the last inequality follows from $d, n_m \geq 2$.
 Therefore, $\|\mathcal{X}^s\|_\infty\leq \|\mathcal{X}^*\|_\infty+\frac{c}{2}\leq c$, which implies $\mathcal{X}^s\in \Upsilon$.
As a consequence, we get that
\begin{equation}\label{KLMSXGN}
\begin{split}
\min_{\mathcal{X}\in\Upsilon}\Biggl\{ \frac{K(p_{\mathcal{X}^*}(\mathcal{Y})|| p_{\mathcal{X}}(\mathcal{Y}))}{n_1n_2\cdots n_d}\Biggl\}
& =
\min_{\mathcal{X}\in\Upsilon}\Biggl\{ \frac{\|\mathcal{X}-\mathcal{X}^*\|_F^2}{2\sigma^2n_1n_2\cdots n_d}\Biggl\} \\
& \leq \frac{\|\mathcal{X}^s-\mathcal{X}^*\|_F^2}{2\sigma^2n_1n_2\cdots n_d}  \leq \frac{\|\mathcal{X}^s-\mathcal{X}^*\|_\infty^2}{2\sigma^2} \\
& \leq \frac{c^2}{2\sigma^2dn_m}\leq   \frac{c^2}{2\sigma^2m},
\end{split}
\end{equation}
where the third inequality holds by (\ref{taulowb}) and Lemma \ref{InfNor},
and the last inequality holds by the fact that  $m\leq dn_m$.

Moreover, by the construction of $A_i^s$ in Lemma \ref{InfNor},
we know that $\|A_i^s\|_0=\|A_i^*\|_0, \forall i\in [d]$.
Therefore, plugging (\ref{KLMSXGN}) and (\ref{lambdai}) into (\ref{ChaTheGa}), we have
\[
\begin{split}
  \frac{\mathbb{E}_{\Omega,\mathcal{Y}_{\Omega}}\left[\|\mathcal{X}^\lambda-\mathcal{X}^*\|_F^2\right]}{n_1n_2\cdots n_d}
\leq & \ \frac{22c^2\log(m)}{m} \\
&  \ + 16(\beta+2)(2c^2+3\sigma^2)\left(\frac{r_1r_2\cdots r_d+\sum_{i=1}^d\|A_i^*\|_0}{m}\right)\log(n_m),
\end{split}
\]
which leads to the desired conclusion.  \qed

\section*{Appendix C. Proof of Theorem \ref{errLaplacNoi}}

{\bf Proof.} For the observations with additive  Laplace noise,
by \cite[Lemma 3.1]{sparse2020zhang}, we know that
\[
\begin{split}
& \  K\left(p_{\mathcal{X}_{i_1i_2\cdots i_d}^*}(\mathcal{Y}_{i_1i_2\cdots i_d})||p_{\mathcal{X}_{i_1i_2\cdots i_d}}(\mathcal{Y}_{i_1i_2\cdots i_d})\right)\\
=& \ \frac{|\mathcal{X}_{i_1i_2\cdots i_d}-\mathcal{X}_{i_1i_2\cdots i_d}^*|}{\tau}-1+\exp\left(-\frac{|\mathcal{X}_{i_1i_2\cdots i_d}-\mathcal{X}_{i_1i_2\cdots i_d}^*|}{\tau}\right) \\
\leq & \ \frac{(\mathcal{X}_{i_1i_2\cdots i_d}-\mathcal{X}_{i_1i_2\cdots i_d}^*)^2}{2\tau^2},
\end{split}
\]
where the inequality follows from the fact that $e^{-x}\leq 1-x+\frac{x^2}{2}$ for any $x>0$.
Therefore, we can choose $\gamma=\frac{c^2}{2\tau^2}$ in (\ref{kappar}).

Let $f(t):=\log(1+\frac{t}{2\tau})$, where $t\in[0,c]$.
Then by  Taylor's expansion, we get that
\begin{equation}\label{TEFX}
f(t)=f(0)+f'(0)t+\frac{f''(\xi)}{2}t^2,
\end{equation}
 where $\xi\in[0,c]$.
By \cite[Lemma 3.1]{sparse2020zhang}, we obtain that
\[
\begin{split}
& \ -2\log(H(p_{\mathcal{X}_{i_1i_2\cdots i_d}}
(\mathcal{Y}_{i_1i_2\cdots i_d}),p_{(\mathcal{X}^*)_{i_1i_2\cdots i_d}}(\mathcal{Y}_{i_1i_2\cdots i_d}))) \\
= & \  \frac{|\mathcal{X}_{i_1i_2\cdots i_d}-\mathcal{X}_{i_1i_2\cdots i_d}^*|}{\tau} - 2\log\left(1+\frac{|\mathcal{X}_{i_1i_2\cdots i_d}-\mathcal{X}_{i_1i_2\cdots i_d}^*|}{2\tau}\right) \\
\geq & \ \frac{(\mathcal{X}_{i_1i_2\cdots i_d}-\mathcal{X}_{i_1i_2\cdots i_d}^*)^2}{(2\tau+c)^2}
\end{split}
\]
where the inequality holds by letting $t = |\mathcal{X}_{i_1i_2\cdots i_d}-\mathcal{X}_{i_1i_2\cdots i_d}^*|$ and the fact that
$f''(\xi)\leq -\frac{1}{(2\tau+c)^2}$ in (\ref{TEFX}).
Therefore, by Theorem \ref{mainthe}, we can deduce
\begin{equation}\label{LapNEEb}
\begin{split}
& \  \frac{\mathbb{E}_{\Omega,\mathcal{Y}_{\Omega}}\left[\|\mathcal{X}^\lambda-\mathcal{X}^*\|_F^2\right]}{n_1n_2\cdots n_d}
\leq \frac{8c^2(2\tau+c)^2\log(m)}{2\tau^2m} + 3(2\tau+c)^2\cdot\min_{\mathcal{X}\in\Upsilon}\Biggl\{ \frac{K(p_{\mathcal{X}^*}(\mathcal{Y})|| p_{\mathcal{X}}(\mathcal{Y}))}{n_1n_2\cdots n_d}\\
& \ ~~~~~~~~~~~~~~~~~~~~~~~~~~~~~~~~~~ +\left(\max_i\{\lambda_i\}+\frac{4c^2(\beta+2)\log(n_m)}{3\tau^2}\right)\frac{r_1r_2\cdots r_d+\sum_{i=1}^d\|A_i\|_0}{m} \Biggl\}.
\end{split}
\end{equation}
Then by a similar argument of (\ref{KLMSXGN}) in the proof of Theorem \ref{errspeGaNo},
we know that
\begin{equation}\label{LapMKL}
\begin{split}
\min_{\mathcal{X}\in\Upsilon}\Biggl\{ \frac{K(p_{\mathcal{X}^*}(\mathcal{Y})|| p_{\mathcal{X}}(\mathcal{Y}))}{n_1n_2\cdots n_d}\Biggl\}
& \leq
\min_{\mathcal{X}\in\Upsilon}\Biggl\{ \frac{\|\mathcal{X}-\mathcal{X}^*\|_F^2}{2\tau^2n_1n_2\cdots n_d}\Biggl\} \\
& \leq \frac{\|\mathcal{X}^s-\mathcal{X}^*\|_F^2}{2\tau^2n_1n_2\cdots n_d}  \leq \frac{\|\mathcal{X}^s-\mathcal{X}^*\|_\infty^2}{2\tau^2}
\leq   \frac{c^2}{2\tau^2m},
\end{split}
\end{equation}
where $\mathcal{X}^s$ is defined in Lemma \ref{InfNor}.
% and the last inequality holds by (\ref{taulowb}), Lemma \ref{InfNor},
%and the fact that  $m\leq dn_m$.
By the construction of $A_i^s$ in Lemma \ref{InfNor},
we know that $\|A_i^s\|_0=\|A_i^*\|_0, \forall i\in [d]$.
As a consequence, plugging (\ref{LapMKL}) and (\ref{lambdai}) into (\ref{LapNEEb}),
the estimator of (\ref{model}) satisfies
\[
\begin{split}
  \frac{\mathbb{E}_{\Omega,\mathcal{Y}_{\Omega}}  \left[|\mathcal{X}^{\lambda}-\mathcal{X}^*\|_F^2\right]}{n_1n_2\cdots n_d}
 \leq & \  \frac{11c^2(2\tau+c)^2\log(m)}{2\tau^2m} \\
 & \ + 12\left(1+\frac{2c^2}{3\tau^2}\right)(2\tau+c)^2(\beta+2)\log(n_m)\frac{r_1r_2\cdots r_d+\sum_{i=1}^d\|A_i^*\|_0}{m},
\end{split}
\]
which is the desired statement. \qed

\section*{Appendix D. Proof of Theorem \ref{poissoerrb}}

{\bf Proof.}
By \cite[Lemma 8]{cao2016poisson}, we obtain that the KL divergence of Poisson observations is
\[
\begin{split}
K\left(p_{\mathcal{X}_{i_1i_2\cdots i_d}^*}(\mathcal{Y}_{i_1i_2\cdots i_d})||p_{\mathcal{X}_{i_1i_2\cdots i_d}}(\mathcal{Y}_{i_1i_2\cdots i_d})\right)
& \leq \frac{1}{\mathcal{X}_{i_1i_2\cdots i_d}}(\mathcal{X}_{i_1i_2\cdots i_d}-\mathcal{X}_{i_1i_2\cdots i_d}^*)^2 \\
& \leq \frac{1}{\varrho}(\mathcal{X}_{i_1i_2\cdots i_d}-\mathcal{X}_{i_1i_2\cdots i_d}^*)^2.
\end{split}
\]
Consequently, we can choose $\gamma=  \frac{c^2}{\varrho}$ in (\ref{lambdai}).
Moreover, according to the proof of \cite[Appendix IV]{raginsky2010compressed}, we have
\[
\begin{split}
-2\log(H(p_{\mathcal{X}_{i_1i_2\cdots i_d}}
(\mathcal{Y}_{i_1i_2\cdots i_d}),p_{(\mathcal{X}^*)_{i_1i_2\cdots i_d}}(\mathcal{Y}_{i_1i_2\cdots i_d})))
& = \left(\sqrt{\mathcal{X}_{i_1i_2\cdots i_d}^*}-\sqrt{\mathcal{X}_{i_1i_2\cdots i_d}}\right)^2 \\
& \geq \frac{1}{4c}\left(\mathcal{X}_{i_1i_2\cdots i_d}-\mathcal{X}_{i_1i_2\cdots i_d}^*\right)^2,
\end{split}
\]
where the inequality holds by
\[
\begin{split}
(\mathcal{X}_{i_1i_2\cdots i_d}-\mathcal{X}_{i_1i_2\cdots i_d}^*)^2 & =\left(\left(\sqrt{\mathcal{X}_{i_1i_2\cdots i_d}^*}-\sqrt{\mathcal{X}_{i_1i_2\cdots i_d}}\right)\left(\sqrt{\mathcal{X}_{i_1i_2\cdots i_d}^*}+\sqrt{\mathcal{X}_{i_1i_2\cdots i_d}}\right)\right)^2 \\
&  \leq 4c\left(\sqrt{\mathcal{X}_{i_1i_2\cdots i_d}^*}-\sqrt{\mathcal{X}_{i_1i_2\cdots i_d}}\right)^2.
\end{split}
\]
Hence, by Theorem \ref{mainthe}, we get that
\begin{equation}\label{PoiEBTheGa}
\begin{split}
& \  \frac{\mathbb{E}_{\Omega,\mathcal{Y}_{\Omega}}\left[\|\mathcal{X}^\lambda-\mathcal{X}^*\|_F^2\right]}{n_1n_2\cdots n_d}
\leq \frac{32c^3\log(m)}{m\varrho}+ 12c\cdot\min_{\mathcal{X}\in\Upsilon}\Biggl\{ \frac{K(p_{\mathcal{X}^*}(\mathcal{Y})|| p_{\mathcal{X}}(\mathcal{Y}))}{n_1n_2\cdots n_d}\\
& \ ~~~~~~~~~~~~~~~~~~~~~~~~~~~~~~~~~~ +\left(\max_i\{\lambda_i\}+\frac{8c^2(\beta+2)\log(n_m)}{3\varrho}\right)\frac{r_1r_2\cdots r_d+\sum_{i=1}^d\|A_i\|_0}{m} \Biggl\}.
\end{split}
\end{equation}
Let $\mathcal{X}^s$ be defined as in Lemma \ref{InfNor}.
Then by a similar argument of (\ref{KLMSXGN}), we get that
\begin{equation}\label{PoiKLMSX}
\begin{split}
\min_{\mathcal{X}\in\Upsilon}\Biggl\{ \frac{K(p_{\mathcal{X}^*}(\mathcal{Y})|| p_{\mathcal{X}}(\mathcal{Y}))}{n_1n_2\cdots n_d}\Biggl\}
& \leq
\min_{\mathcal{X}\in\Upsilon}\Biggl\{ \frac{\|\mathcal{X}-\mathcal{X}^*\|_F^2}{\varrho n_1n_2\cdots n_d}\Biggl\} \\
& \leq \frac{\|\mathcal{X}^s-\mathcal{X}^*\|_F^2}{\varrho n_1n_2\cdots n_d}  \leq \frac{\|\mathcal{X}^s-\mathcal{X}^*\|_\infty^2}{\varrho} \leq   \frac{c^2}{\varrho m}.
\end{split}
\end{equation}
%where the third inequality holds by (\ref{taulowb}) and the last inequality holds by $m\leq dn_m$.
Notice that $\|A_i^s\|_0=\|A_i^*\|_0, \forall i\in[d]$.
By combining (\ref{lambdai}), (\ref{PoiEBTheGa}), and (\ref{PoiKLMSX}), after some rearrangements,
we obtain that  the estimator of (\ref{model}) satisfies
\[
\begin{split}
  \frac{\mathbb{E}_{\Omega,\mathcal{Y}_{\Omega}}  \left[|\mathcal{X}^{\lambda}-\mathcal{X}^*\|_F^2\right]}{n_1n_2\cdots n_d}
 \leq & \  \frac{44c^3\log(m)}{\varrho m}  + 48c\left(1+\frac{4c^2}{3\varrho}\right)(\beta+2)\log(n_m)\frac{r_1r_2\cdots r_d+\sum_{i=1}^d\|A_i^*\|_0}{m}.
\end{split}
\]
This concludes the proof. \qed

\section*{Appendix E. Proof of Theorem \ref{MiniMaxLowbg}}

Let
$$
 \mathfrak{L}= \left\{\mathcal{X} = \mathcal{C}\times_1A_1\cdots\times_d A_d: \mathcal{C}\in\mathfrak{D}, A_i\in \mathfrak{X}_i, i\in[d]\right\},
$$
where
$$
\mathfrak{D}=\left\{\mathcal{C}\in\mathbb{R}_+^{r_1\times r_2\times\cdots\times r_d}:\mathcal{C}_{i_1\cdots i_d}\in\{0,1,c_0\}, (i_1,\ldots, i_d)\in[n_1]\times \cdots \times [n_d]\right\}
$$
with
\begin{equation}\label{CoOD}
 c_0= \min\left\{1,\frac{\gamma_c\mu}{\prod_{i=1}^d(a_i\sqrt{\Delta_i(s_i,n_i)})}\sqrt{\frac{r_1r_2\cdots r_d}{m}}\right\},
\end{equation}
and
$$
\mathfrak{X}_i=\{A_i\in\mathbb{R}_+^{n_i\times r_i}:(A_i)_{jk}\in \{0,a_i,  b_i\}, \|A_i\|_0\leq s_i, (j,k)\in[n_i]\times [r_i]\}
$$
with
$$
b_i=\min\left\{a_i, \frac{\gamma_i\mu}{\left(\prod_{j\neq i}a_j\right)\sqrt{\prod_{j=1}^d\Delta(s_j,n_j)}}\sqrt{\frac{s_i}{m}}\right\}, \ i\in[d].
$$
We will specify $\gamma_c, \gamma_i, i\in[d]$ in detail later.
Therefore, by the construction of $ \mathfrak{L}$,
 we know that $ \mathfrak{L}\subseteq  \mathfrak{L}(\mathbf{s},\mathbf{r},\mathbf{a})$.
 Next we consider to construct the packing sets of the core tensor and factor matrices, respectively.

\emph{Case I}. We construct the following set
$$
\mathfrak{L}_\mathcal{C}=\{\mathcal{X}= \mathcal{C}\times_1A_1\cdots\times_d A_d: \mathcal{C}\in \mathfrak{D}_0\},
$$
where $\mathfrak{D}_0$ is defined as
\begin{equation}\label{DOC}
\mathfrak{D}_0=\left\{\mathcal{C}\in\mathbb{R}_+^{r_1\times r_2\times\cdots\times r_d}:
\mathcal{C}_{i_1i_2\cdots i_d}\in\{0,c_0\}, (i_1,\ldots, i_d)\in[n_1]\times \cdots \times [n_d]\right\}
\end{equation}
and
$$
A_i=a_i\begin{pmatrix} I_{r_i} \\ \vdots \\ I_{r_i} \\  \mathbf{0}_{r_i} \end{pmatrix}\in\mathbb{R}_+^{n_i\times r_i} \ \textup{with} \ I_{r_i}\in\mathbb{R}^{r_i\times r_i}, \
 i\in[d].
$$
There are $\lfloor\frac{s_i\wedge n_i}{r_i}\rfloor$ blocks identity matrix $I_{r_i}$ in $A_i$ and $\mathbf{0}_{r_i}\in \mathbb{R}^{(n_i-r_i\lfloor\frac{s_i\wedge n_i}{r_i}\rfloor)\times r_i}$ is a zero matrix.
Consequently, $\mathfrak{L}_\mathcal{C}\subseteq  \mathfrak{L}$.

For any $\mathcal{X}\in\mathfrak{L}_\mathcal{C}$, by (\ref{TenUnfol}),
 we have
\[
\begin{split}
\mathcal{X}_{(1)}&= A_1\mathcal{C}_{(1)}(A_d\otimes A_{d-1}\otimes\cdots \otimes A_2)^T \\
%% & = a_1\begin{pmatrix} I_{r_1} \\ \vdots \\ I_{r_1} \\  \mathbf{0}_{r_1} \end{pmatrix}\mathcal{C}_{(1)}(A_2\otimes A_3\otimes\cdots \otimes A_d)^T \\
& = a_1\begin{pmatrix} \mathcal{C}_{(1)} \\ \vdots \\ \mathcal{C}_{(1)} \\  \mathbf{0}_{r_{12}} \end{pmatrix}(A_d\otimes A_{d-1}\otimes\cdots \otimes A_2)^T
\end{split}
\]
where $\mathbf{0}_{r_{12}}\in\mathbb{R}^{(n_1-r_1\lfloor\frac{s_1\wedge n_1}{r_1}\rfloor)\times (r_2\cdots r_d)}$ is a zero matrix.
 By some simple calculations, we get
\begin{equation}\label{NFXT}
\|\mathcal{X}\|_F^2=\|\mathcal{X}_{(1)}\|_F^2=(a_1\cdots a_d)^2\|\mathcal{C}_{(1)}\|_F^2\prod_{i=1}^d\left\lfloor\frac{s_i\wedge n_i}{r_i}\right\rfloor.
\end{equation}
Note that the entries of $\mathcal{C}$ only takes $0$ or $c_0$ by the construction of $\mathfrak{D}_0$ in (\ref{DOC}).
By the Varshamov-Gilbert bound \cite[Lemma 2.9]{tsybakov2009},
we know that there exists a subset $\mathfrak{L}_\mathcal{C}^0\subseteq \mathfrak{L}_\mathcal{C}$ such that
\begin{equation}\label{CarLo0}
|\mathfrak{L}_\mathcal{C}^0|\geq 2^{r_1\cdots r_d/8}+1,
\end{equation}
and
for any $\mathcal{X}_1,\mathcal{X}_2\in \mathfrak{L}_\mathcal{C}^0$,
\[
\begin{split}
\|\mathcal{X}_1-\mathcal{X}_2\|_F^2
&= \|(\mathcal{X}_1)_{(1)}-(\mathcal{X}_2)_{(1)}\|_F^2 \\
&\geq \frac{r_1\cdots r_d}{8}\left(\prod_{i=1}^d\left\lfloor\frac{s_i\wedge n_i}{r_i}\right\rfloor\right)(c_0a_1\cdots a_d)^2 \\
&\geq \frac{n_1n_2\cdots n_d}{2^{d+3}}\left(\prod_{i=1}^d\Delta_i(s_i,n_i)\right)(c_0a_1\cdots a_d)^2 \\
&= \frac{n_1n_2\cdots n_d}{2^{d+3}}\min\left\{\prod_{i=1}^da_i^2\Delta_i(s_i,n_i),\gamma_c^2\mu^2\frac{r_1r_2\cdots r_d}{m}\right\},
\end{split}
\]
where the second inequality follows from the fact that
$\lfloor x\rfloor\geq\frac{x}{2} $ for
any $x\geq 1$, $\Delta_i(s_i,n_i)$ is defined as (\ref{DSJN}),
and the last equality follows from the definition of $c_0$ in (\ref{CoOD}).

For an arbitrary tensor $\mathcal{X}\in\mathfrak{L}_\mathcal{C}^0$, we have that
\begin{equation}\label{KL0IC}
\begin{split}
K(\mathbb{P}_\mathcal{X},\mathbb{P}_\mathbf{0})& \leq \frac{m}{n_1n_2\cdots n_d}\left(\frac{1}{2\mu^2}\right)\sum_{i_1,i_2,\ldots, i_d}|\mathcal{X}_{i_1i_2\cdots i_d}|^2 \\
& \leq \frac{m}{n_1n_2\cdots n_d}\left(\frac{1}{2\mu^2}\right) c_0^2\left(\prod_{i=1}^d r_ia_i^2\left\lfloor\frac{s_i\wedge n_i}{r_i}\right\rfloor\right)\\
& \leq  \frac{m}{2\mu^2}\min\left\{\prod_{i=1}^d a_i^2\Delta_i(s_i,n_i),\gamma_c^2\mu^2\frac{r_1r_2\cdots r_d}{m}\right\} \\
&\leq \frac{\gamma_c^2r_1\cdots r_d}{2} \leq 4\gamma_c^2\log_2(|\mathfrak{L}_\mathcal{C}^0|-1),
\end{split}
\end{equation}
where the first inequality holds by (\ref{KLON})
and $\Omega\sim  \text{Bern}(p)$ with $p=\frac{m}{n_1n_2\cdots n_d}$,
the second inequality holds by (\ref{NFXT}) and $\mathfrak{L}_\mathcal{C}^0\subseteq \mathfrak{L}_\mathcal{C}$,
 the third inequality holds by $\lfloor x \rfloor\leq x$ for any $x>0$
and the definition of $c_0$ in (\ref{CoOD}),
and the last inequality holds by (\ref{CarLo0}).
Consequently,  summing up the inequality in (\ref{KL0IC})
for all $\mathcal{X}\in\mathfrak{L}_\mathcal{C}^0$
except for $\mathcal{X}=\mathbf{0}$, we deduce
$$
\frac{1}{|\mathfrak{L}_\mathcal{C}^0|-1}\sum_{\mathcal{X}\in\mathfrak{L}_\mathcal{C}^0}K(\mathbb{P}_\mathcal{X},\mathbb{P}_\mathbf{0})
 \leq  \alpha \log(|\mathfrak{L}_\mathcal{C}^0|-1)
$$
where $\gamma_c:=\frac{\sqrt{\alpha\log(2)}}{2}$ with $\alpha\in(0,1/8)$.
Hence, by \cite[Theorem 2.5]{tsybakov2009}, we have
\begin{equation}\label{CLIowb}
\begin{split}
& \ \inf_{\widetilde{\mathcal{X}}}\sup_{\mathcal{X}^*\in \mathfrak{L}(\mathbf{s},\mathbf{r},\mathbf{a})}\mathbb{P}\left(\frac{\|\widetilde{\mathcal{X}}-\mathcal{X}^*\|_F^2}{n_1\cdots n_d}\geq \frac{1}{2^{d+4}}\min\left\{\prod_{i=1}^da_i^2\Delta_i(s_i,n_i),\gamma_c^2\mu^2\frac{r_1r_2\cdots r_d}{m}\right\}\right) \\
\geq & \  \inf_{\widetilde{\mathcal{X}}}\sup_{\mathcal{X}^*\in \mathfrak{L}_{\mathcal{C}}^0}\mathbb{P}\left(\frac{\|\widetilde{\mathcal{X}}-\mathcal{X}^*\|_F^2}{n_1\cdots n_d}\geq \frac{1}{2^{d+4}}\min\left\{\prod_{i=1}^da_i^2\Delta_i(s_i,n_i),\gamma_c^2\mu^2\frac{r_1r_2\cdots r_d}{m}\right\}\right) \\
\geq   & \ \widetilde{\alpha},
\end{split}
\end{equation}
where $\widetilde{\alpha}$ is defined as
$$
 \widetilde{\alpha}:=\frac{\sqrt{|\mathfrak{L}_\mathcal{C}^0|-1}}{1
+\sqrt{|\mathfrak{L}_\mathcal{C}^0|-1}}\left(1-2\alpha-\sqrt{\frac{2\alpha}{\log(|\mathfrak{L}_\mathcal{C}^0|-1)}}\right).
$$

\emph{Case II}. Now we consider the packing set about the factor matrices.
For a fixed $i$,
then for any $j\neq i$, we let
$$
A_j=a_j\begin{pmatrix} I_{r_j} \\ \vdots \\ I_{r_j} \\  \mathbf{0}_j \end{pmatrix}\in\mathbb{R}_+^{n_j\times r_j},
$$
where there are $\lfloor\frac{s_j\wedge n_j}{r_j}\rfloor$ blocks identity matrix $I_{r_j}$ in $A_j$
and $\mathbf{0}_j$ is an $\big(n_j-r_j\lfloor\frac{s_j\wedge n_j}{r_j}\rfloor\big)\times r_j$ zero matrix.
Let
$$
\widetilde{\mathbf{A}}_i: = \left\{A_i= (A_{r_i'} \ \mathbf{0}_{r_i'})\in\mathbb{R}_+^{n_i\times r_i}: A_{r_i'}\in\mathbb{R}_+^{n_i\times r_i'}
\ \textup{with} \ (A_{r_i'})_{kj}\in\{0,b_i\}, \mathbf{0}_{r_i'}\in\mathbb{R}^{n_i\times (r_i-r_i')} \right\},
$$
where $r_i':=\lceil\frac{s_i}{n_i}\rceil$, there are at most $s_i$ nonzero entries in $A_{r_i'}$,
and $\mathbf{0}_{r_i'}$ is an $n_i\times (r_i-r_i')$ zero matrix.
Now we construct the following set
$$
\mathfrak{L}_{A_i}=\left\{\mathcal{X}= \mathcal{C}\times_1A_1\cdots \times_{i-1}A_{i-1}\times_i A_i \times_{i+1} A_{i+1}\cdots \times_d A_d: A_i\in \widetilde{\mathbf{A}}_i\right\},
$$
where the mode-$i$ unfolding of $\mathcal{C}$ is defined as
%where
%$$
%\widetilde{\mathbf{A}}_i = \left\{A_i= (A_{r_i'} \ \mathbf{0}_{r_i'})\in\mathbb{R}_+^{n_i\times r_i}: A_{r_i'}\in\mathbb{R}_+^{n_i\times r_i'}
%\ \textup{with} \ (A_{r_i'})_{kj}\in\{0,b_i\}, \mathbf{0}_{r_i'}\in\mathbb{R}^{n_i\times (r_i-r_i')} \right\},
%$$
%with $r_i'=\lceil\frac{s_i}{n_i}\rceil$, there are at most $s_i$ nonzero entries in $A_{r_i'}$,
$$
\mathcal{C}_{(i)}=
\begin{pmatrix} I_{r_i'}  \ \ \ \  I_{r_i'}             \  \ \ \ \cdots \ \ \ \ I_{r_i'}  \ \ \ \ \  \ \ \ \mathbf{0}_{1r_{i}'} \\
\mathbf{0}_{2r_{i}'}     \ \ \  \  \mathbf{0}_{2r_{i}'}  \ \ \ \  \cdots \ \ \ \ \mathbf{0}_{2r_{i}'} \ \ \  \ \mathbf{0}_{3r_{i}'}
\end{pmatrix}.
$$
Here there are $\lfloor\frac{\prod_{j\neq i}r_j}{r_i'}\rfloor$ blocks identity matrices  $I_{r_i'}\in\mathbb{R}^{r_i'\times r_i'}$ in $\mathcal{C}_{(i)}$,
and
$$\mathbf{0}_{1r_{i}'}\in\mathbb{R}^{r_i'\times \big(\prod_{j\neq i}r_j-r_i'\lfloor\frac{\prod_{j\neq i}r_j}{r_i'}\rfloor\big)}, \ \mathbf{0}_{2r_{i}'}\in\mathbb{R}^{(r_i-r_i')\times r_i' }, \
\mathbf{0}_{3r_{i}'}\in \mathbb{R}^{(r_i-r_i')\times \big(\prod_{j\neq i}r_j-r_i'\lfloor\frac{\prod_{j\neq i}r_j}{r_i'}\rfloor\big)}
$$
are zero matrices.
From this construction, we know that $\mathfrak{L}_{A_i}\subseteq  \mathfrak{L}$.
Therefore, for any $\mathcal{X}\in\mathfrak{L}_{A_i}$, we have
\[
\begin{split}
\mathcal{X}_{(i)}&= A_i\mathcal{C}_{(i)}(A_{d}\otimes\cdots \otimes A_{i+1}\otimes A_{i-1} \otimes\cdots \otimes A_{1})^T \\
& = (A_{r_i'} \  \cdots \ A_{r_i'} \ \mathbf{0}_{n_i} )(A_{d}\otimes\cdots \otimes A_{i+1}\otimes A_{i-1} \otimes\cdots \otimes A_{1})^T,
\end{split}
\]
where the first equality follows from (\ref{TenUnfol}) and
 $\mathbf{0}_{n_i}\in\mathbb{R}^{n_i\times \big(\prod_{j\neq i}r_j-r_i'\lfloor\frac{\prod_{j\neq i}r_j}{r_i'}\rfloor\big)}$ is a zero matrix.
Hence,  we can obtain through some simple calculations that
\begin{equation}\label{FmXNF}
\|\mathcal{X}\|_F^2=\|\mathcal{X}_{(i)}\|_F^2=\left\lfloor\frac{\prod_{j\neq i}r_j}{r_i'}\right\rfloor\left(\prod_{j\neq i}\left(\left\lfloor\frac{s_j\wedge n_j}{r_j}\right\rfloor a_j^2\right)\right)\|A_{r_i'}\|_F^2.
\end{equation}
Notice  that the entries of $A_{r_i'}$ only takes $0$ or $b_i$.
By the Varshamov-Gilbert bound \cite[Lemma 2.9]{tsybakov2009},
we know that there exists $\mathfrak{L}_{A_i}^0\subseteq \mathfrak{L}_{A_i}$ such that
\begin{equation}\label{LAiC}
|\mathfrak{L}_{A_i}^0|\geq 2^{s_i/8}+1,
\end{equation}
 and
for any $\mathcal{X}_1,\mathcal{X}_2\in \mathfrak{L}_{A_i}^0$,
\begin{equation}
\begin{split}
\|\mathcal{X}_1-\mathcal{X}_2\|_F^2
&\geq \frac{s_i}{8}\left\lfloor\frac{\prod_{j\neq i}r_j}{r_i'}\right\rfloor\prod_{j\neq i}\left(\left\lfloor\frac{s_j\wedge n_j}{r_j}\right\rfloor a_j^2\right) b_i^2 \\
&\geq \frac{s_i b_i^2}{2^{d+3}r_{i}'} \prod_{j\neq i}\left({s_j\wedge n_j}\right)a_j^2\\
&\geq \frac{n_1\cdots n_d}{2^{d+4}}\left(\prod_{j=1}^d \Delta_j(s_j,n_j)\right)\left(\prod_{j\neq i}a_j^2\right)\min\left\{a_i^2, \frac{\gamma_i^2\mu^2}{(\prod_{j\neq i}a_j^2)\prod_{j=1}^d\Delta_j(s_j,n_j)}\left(\frac{s_i}{m}\right)\right\} \\
&=\frac{n_1\cdots n_d}{2^{d+4}}\min\left\{\prod_{j=1}^d\Delta_j(s_j,n_j)a_j^2,\gamma_i^2\mu^2\left(\frac{s_i}{m}\right) \right\},
\end{split}
\end{equation}
where the second inequality holds by  $\lfloor x\rfloor\geq\frac{x}{2} $ for any $x\geq 1$
 and the third inequality holds by $x/\lceil x  \rceil\geq \frac{1}{2}\min\{x,1\}$ for any $x>0$.

In addition, for any $\mathcal{X}\in \mathfrak{L}_{A_i}^0$, we have that
\begin{equation}\label{KLoAi}
\begin{split}
K(\mathbb{P}_\mathcal{X},\mathbb{P}_\mathbf{0})
& \leq \frac{m}{n_1n_2\cdots n_d}\left(\frac{1}{2\mu^2}\right)\sum_{i_1,i_2,\ldots, i_d}|\mathcal{X}_{i_1i_2\cdots i_d}|^2 \\
& \leq \frac{m}{n_1n_2\cdots n_d}\left(\frac{1}{2\mu^2}\right) \left\lfloor\frac{\prod_{j\neq i}r_j}{r_i'}\right\rfloor\prod_{j\neq i}\left(\left\lfloor\frac{s_j\wedge n_j}{r_j}\right\rfloor a_j^2\right)(n_ir_i'\wedge s_i)b_i^2\\
&\leq  \frac{m}{2\mu^2} \left(\prod_{j=1}^d \Delta_j(s_j,n_j)\right) \left(\prod_{j\neq i} a_j\right)b_i^2\\
& =  \frac{m}{2\mu^2}\min\left\{\prod_{j=1}^d a_j^2\Delta_j(s_j,n_j),\gamma_i^2\mu^2\frac{s_i}{m}\right\} \\
&\leq \frac{\gamma_i^2s_i}{2} \leq 4\gamma_i^2\log_2(|\mathfrak{L}_{A_i}^0|-1),
\end{split}
\end{equation}
where the first inequality holds by (\ref{KLON})
and $\Omega\sim  \text{Bern}(p)$ with $p=\frac{m}{n_1n_2\cdots n_d}$,
the second inequality holds by (\ref{FmXNF}) and $\|A_{r_i'}\|_F^2\leq (n_ir_i'\wedge s_i)b_i^2$,
the third inequality holds by the fact that $\lfloor x \rfloor\leq x$ for any $x>0$,
 and the last inequality holds by (\ref{LAiC}).
Let  $\gamma_i={\sqrt{\alpha_i\log(2)}}/{2}$ with $\alpha_i\in(0,\frac{1}{8})$.
Summing up the inequality in (\ref{KLoAi})
for all $0\neq\mathcal{X}\in \mathfrak{L}_{A_i}$ yields
$$
\frac{1}{|\mathfrak{L}_{A_i}^0|-1}\sum_{\mathcal{X}\in\mathfrak{L}_{A_i}^0}K(\mathbb{P}_\mathcal{X},\mathbb{P}_\mathbf{0}) \leq \alpha_i\log(|\mathfrak{L}_{A_i}^0|-1).
$$
It follows from \cite[Theorem 2.5]{tsybakov2009} that
\begin{equation}\label{AiLowbd}
\begin{split}
& \  \inf_{\widetilde{\mathcal{X}}}\sup_{\mathcal{X}^*\in \mathfrak{L}(\mathbf{s},\mathbf{r},\mathbf{a})}\mathbb{P}\left(\frac{\|\widetilde{\mathcal{X}}-\mathcal{X}^*\|_F^2}{n_1\cdots n_d}\geq \frac{1}{2^{d+5}}\min\left\{\prod_{j=1}^d\Delta_j(s_j,n_j)a_j^2,\gamma_i^2\mu^2\left(\frac{s_i}{m}\right) \right\}\right) \\
\geq & \ \inf_{\widetilde{\mathcal{X}}}\sup_{\mathcal{X}^*\in \mathfrak{L}_{A_i}^0}\mathbb{P}\left(\frac{\|\widetilde{\mathcal{X}}-\mathcal{X}^*\|_F^2}{n_1\cdots n_d}\geq \frac{1}{2^{d+5}}\min\left\{\prod_{j=1}^d\Delta_j(s_j,n_j)a_j^2,\gamma_i^2\mu^2\left(\frac{s_i}{m}\right) \right\}\right) \\
\geq   & \ \widetilde{\alpha}_i,
\end{split}
\end{equation}
where
$$
\widetilde{\alpha}_i=\frac{\sqrt{|\mathfrak{L}_{A_i}^0|-1}}{1
+\sqrt{|\mathfrak{L}_{A_i}^0|-1}}\left(1-2\alpha_i-\sqrt{\frac{2\alpha_i}{\log(|\mathfrak{L}_{A_i}^0|-1)}}\right).
$$

Let $$
\eta:=\min\left\{\prod_{j=1}^d\Delta_j(s_j,n_j)a_j^2,\gamma_m^2\mu^2\left(\frac{r_1r_2\cdots r_d+\sum_{i=1}^d s_i}{m}\right) \right\},
$$
where  $\gamma_m=\min\{\gamma_c,\gamma_1,\ldots, \gamma_d\}$.
Combining (\ref{CLIowb}) with (\ref{AiLowbd}), we deduce that
\[
\begin{split}
&\inf_{\widetilde{\mathcal{X}}}\sup_{\mathcal{X}^*\in \mathfrak{L}(\mathbf{s},\mathbf{r},\mathbf{a})}\mathbb{P}\left(\frac{\|\widetilde{\mathcal{X}}-\mathcal{X}^*\|_F^2}{n_1\cdots n_d}\geq \frac{1}{2^{d+5}(d+1)}\eta \right)
\geq     \widetilde{\alpha}_m,
\end{split}
\]
where
$
\widetilde{\alpha}_m=\min\{\widetilde{\alpha},\widetilde{\alpha}_1,\ldots, \widetilde{\alpha}_d\}\in(0,1).
$
Therefore, by Markov's inequality, we get that
\[
\begin{split}
&\inf_{\widetilde{\mathcal{X}}}\sup_{\mathcal{X}^*\in \mathfrak{L}(\mathbf{s},\mathbf{r},\mathbf{a})}
\frac{\mathbb{E}_{\Omega, \mathcal{Y}_\Omega}\|\widetilde{\mathcal{X}}-\mathcal{X}^*\|_F^2}{n_1\cdots n_d}\\
\geq \ &  \frac{\widetilde{\alpha}_m}{2^{d+5}(d+1)}\min\left\{\prod_{i=1}^d\Delta_i(s_i,n_i)a_i^2,\gamma_m^2\mu^2\left(\frac{r_1r_2\cdots r_d+\sum_{i=1}^d s_i}{m}\right) \right\},
\end{split}
\]
which concludes the proof. \qed

\bibliographystyle{abbrv}
%\bibliographystyle{siam}
%
%plain������ĸ��˳�����У��Ƚϴ���Ϊ���ߡ����Ⱥͱ��⣻
%unsrt����ʽͬplain��ֻ�ǰ������õ��Ⱥ�������
%alpha��������������ĸ+���ݺ���λ�����ţ�����ĸ˳��������
%abbrv������plain�����·�ȫƴ��Ϊ��д�����Խ��գ�
%ieeetr�����ʵ������ӹ���ʦЭ���ڿ���ʽ��
%acm������������ѧ���ڿ���ʽ��
%siam��������ҵ��Ӧ����ѧѧ���ڿ���ʽ��
%apalike����������ѧѧ���ڿ���ʽ��

\bibliography{RefPSparsTuc}

\end{document}